\documentclass{article}

\usepackage{microtype}
\usepackage{graphicx}
\usepackage{subfigure}
\usepackage{booktabs} 

\usepackage{hyperref}

\usepackage[accepted]{icml2024}

\usepackage{amsmath}
\usepackage{amssymb}
\usepackage{mathtools}
\usepackage{amsthm}
\usepackage{bbm}
\usepackage{multirow}
\usepackage{booktabs}
\usepackage{siunitx}
\usepackage {indentfirst}
\usepackage{float} 

\usepackage[capitalize,noabbrev]{cleveref}

\theoremstyle{plain}
\newtheorem{theorem}{Theorem}[section]
\newtheorem{proposition}[theorem]{Proposition}

\theoremstyle{definition}
\newtheorem{definition}[theorem]{Definition}

\theoremstyle{remark}
\newtheorem{remark}[theorem]{Remark}

\usepackage[textsize=tiny]{todonotes}

\icmltitlerunning{Post-Fair Federated Learning}

\begin{document}

\onecolumn
\icmltitle{Post-Fair Federated Learning: Achieving Group and Community Fairness \\in Federated Learning via Post-processing }




\icmlsetsymbol{equal}{*}

\begin{icmlauthorlist}
\icmlauthor{Yuying Duan}{yyy}
\icmlauthor{Yijun Tian}{yyy}
\icmlauthor{Nitesh Chawla}{yyy}
\icmlauthor{Michael Lemmon}{yyy}
\end{icmlauthorlist}

\let\thefootnote\relax\footnotetext{$^1$ University of Notre Dame,  \{yduan2, ytian5, nchawla, lemmon\}@nd.edu}

\icmlkeywords{Machine Learning, ICML}

\vskip 0.3in




\begin{abstract}
Federated Learning (FL) is a distributed machine learning framework in which a set of local communities collaboratively learn a shared global model while retaining all training data locally within each community.  
Two notions of fairness have recently emerged as important issues for federated learning: group fairness and community fairness.  Group fairness requires that a model's decisions do not favor any particular group based on a set of legally protected attributes such as race or gender.  Community fairness requires that global models exhibit similar levels of performance (accuracy) across all collaborating communities.  Both fairness concepts can coexist within an FL framework, but the existing literature has focused on either one concept or the other.  This paper proposes and analyzes a post-processing \emph{fair federated learning} (FFL) framework called 
 \emph{post-FFL}.  Post-FFL uses a linear program to simultaneously enforce group and community fairness while maximizing the utility of the global model. Because Post-FFL is a post-processing approach, it can be used with existing FL training pipelines whose convergence properties are well understood. This paper uses post-FFL on real-world datasets to mimic how hospital networks, for example, use federated learning to deliver community health care. Theoretical results bound the accuracy lost when post-FFL enforces both notion of fairness. Experimental results illustrate that post-FFL simultaneously improves both group and community fairness in FL. Moreover, post-FFL outperforms the existing in-processing fair federated learning in terms of improving both notions of fairness, communication efficiency and computation cost.
\end{abstract}
\section{Introduction}

Federated Learning (FL) \cite{mcmahan2017communication} is a distributed machine learning framework that uses data collected 
from a group of community \emph{clients} to learn a global model that can be used by all clients in the group.  The communities served by these clients are formed from sets of remote users (e.g. mobile phones) or organizations (e.g. medical clinics and hospitals) that all share some defining attribute such as a similar geographical location. FL algorithms such as \texttt{FedAvg} \cite{mcmahan2017communication} train the global model in a distributed manner by first having each community client use its local data to train a local model.  This local model is then sent to the cloud server who averages these models and sends the averaged model back to the community clients who then retrain that model with their local data.  This interaction between the clients and server continues for a several update cycles until it converges on a global model that is agreeable to all clients.   While there were no theoretical convergence guarantees with the original FL algorithm \cite{mcmahan2017communication}, subsequent analysis \cite{smith2018cocoa,li2020federated} did provide
 theoretical convergence analysis for this FL training pipeline.  Since then 
 FL has come to be a dominant framework for distributed machine learning \cite{kairouz2021advances}, particularly in smart city \cite{jiang2020federated,zheng2022applications,qolomany2020particle,pandya2023federated} and smart healthcare applications \cite{rieke2020future,nguyen2022federated,antunes2022federated,brisimi2018federated} .  
 
 This paper considers two related notions of fairness relevant to federated learning: Group Fairness and Community Fairness.  
 Group fairness \cite{dwork2012fairness,hardt2016equality,zafar2017fairness} is concerned with achieving similar outcomes for groups defined by legally protected (a.k.a. sensitive) attributes such as race or gender. For community fairness \cite{li2019fair, gross2007community, gross2008measure}, a community may consist of individuals living in the same geographic location.    Community fairness, therefore, is more concerned with ensuring that these geographically distinct communities have equal access to resources.  Group fairness, on the other hand, requires that all individuals with the same legally protected attribute receive the same benefits as those outside of the protected group regardless of their membership in these geographically distinct communities or neighborhoods.   
 Both fairness concepts (group vs. community) are relevant to federated learning.  This is particularly true in smart healthcare applications where a physician's decisions should not be influenced by factors such as age, race, or gender  \cite{parsa2017revised} and yet city leaders want to ensure that all geographically distinct neighborhoods perceive they have the same accessibility to adequate health care.  Whether one can balance these two fairness concepts in an FL platform and what might be the cost of attaining such balance is the main topic of this paper.  

This paper defines group fairness with respect to Equal Opportunity (EO) \cite{hardt2016equality} and community fairness with respect to Fair Resource Allocation \cite{li2019fair}. EO ensures that those who deserve services receive them in a manner that is independent of any sensitive attributes. Fair Resource Allocation requires the likelihood of accurate outcomes to be uniform across all participating communities. 
We apply EO to group fairness because most legal standards, such as those from the Equal Employment Opportunity Commission \cite{peck1975equal}, mandate that the true positive rate be consistent across sensitive attributes. However, EO may not be appropriate for communities that receive resources (services or infrastructure) from the city. EO focuses exclusively on individuals who deserve to benefit from resources (true positives), without preventing unqualified individuals from receiving resources. It could lead to surplus resource allocation in some communities.
One example of this scenario might be seen in health care delivery, where a hospital network sees that the number of clinics per capita is equal between rural and urban residents. However, rural residents may still perceive inequality because the effort required to access these clinics (distance traveled) is much greater than that of the urban dwellers.
Fair Resource Allocation considers both true positives and true negatives across all communities. This paper examines whether it is possible to achieve Equal Opportunity and Fair Resources Allocation in federated learning and seeks to estimate the cost we incur in achieving both fairness objectives.

 Several recent papers have proposed methods for achieving  either group or community fairness in federated learning.  The methods fall into three categories: pre-processing, in-processing, and post-processing.  Pre-processing techniques achieve model fairness by modifying the data set used to train the model.  This may be done by weighting the training samples as described in \cite{abay2020mitigating}.  Pre-processing techniques, however, cannot simultaneously address group and community fairness.  In-processing techniques typically modify the federated learning framework's optimization algorithms.  Current approaches either employ dynamic aggregation weights \cite{yue2023gifair,ezzeldin2023fairfed,chu2021fedfair,rodriguez2021enforcing,lyu2020collaborative,li2019fair} or use adversarial training \cite{du2021fairness, mohri2019agnostic}. These approaches, however, complicate the existing FL training pipeline and lack formal convergence guarantees.  Our paper is the first to apply post-processing in FL. Post-processing uses models selected by an existing training pipeline to generate a randomized model that achieves fairness. 
  Prior post-processing work \cite{hardt2016equality,fish2016confidence,menon2017cost,pleiss2017fairness,chzhen2019leveraging,denis2021fairness,zhao2022inherent,zeng2022bayes,xian2023fair} does not consider federated learning (FL), nor does it simultaneously address community and group fairness, which is the subject of this paper.

This paper's novel contributions are: 
\vspace{-0.1in }
\begin{itemize}
\setlength{\parskip}{0.1pt}
\setlength{\parskip}{0.1pt}
    \item The first application of the post-processing technique in federated learning to improve fairness.
    \item Development of a post-processing fair FL framework (post-FFL) that simultaneously enforces group fairness, i.e, Equal Opportunity and community fairness, i.e., Fair Reources Allocation in both binary and multi-class classification.
    \item Results that characterize when post-processing can simultaneously achieve group and community fairness for a given group of communities.
    \item Results that enable the evaluation of the model's accuracy loss when achieving group and community fairness.
    \item Finally, experimental results on a real-world dataset that show our framework outperforms existing baselines in improving both notions of fairness, communication efficiency and computational cost.
\end{itemize}

\section{Preliminary Definitions}
\label{sec:prelim}

This section provides a statistical interpretation of group and community fairness  that allows us to address fairness issues in the federated learning of models that predict outcomes for individuals in a group of communities.   The community group is a collection of geographically distinct human communities.  Each community is a \emph{client} that uses the local data it has on its inhabitants to select a local model that predicts health outcomes for a given inhabitant.  All community clients send their local models to a global \emph{server} who then aggregates the models into a global model.  The resulting global model, however, may not be fair either with respect to group or community notions defined below.  The main problem is to find a way to \emph{transform} the global model into a \emph{fair global model}.  Since this transformation is done after the FL pipeline has selected the global model, this is a post-processing approach to achieving fairness.

\emph{Notational Conventions:} This paper will denote random variables using upper case letters, $X$, and lower case letters, $x$ will denote instances of those random variables.  A random variable's distribution will be denoted as $F_X$ and an instance, $x$,  drawn from that distribution will be denoted as $x \sim F_X$.  
Bold face lower case symbols will be reserved for vectors and
bold face upper case symbols will be reserved for matrices.

To formalize our statistical setup, we first need to define the notion of a \textbf{community group}.

\begin{definition}
\label{def:community-group}
A \textbf{community group} consists of $K$ geographically distinct communities that we formally represent as a tuple of jointly random variables, $D = (X,A,C,Y)$ with probability distribution $F_D: \mathcal{X} \times \mathcal{A} \times
\mathcal{C} \times \mathcal{Y} \rightarrow [0,1]$.    An instance of the community group, $(x,a,c,y)$, is called an \textbf{individual} where $x \in \mathcal{X}$ is the individual's \textbf{private data vector},
$a \in \mathcal{A} = \{0,1\}$ denotes the individual's \textbf{protected sensitive attribute}, and $c \in \mathcal{C} = 
\{1,2, \ldots, K\}$ denotes which \textbf{community} the individual belongs to.   The other value, $y \in \mathcal{Y} = \{0,1\}$ denotes the individual's \emph{qualified outcome}.   
\end{definition}

The variables in definition \ref{def:community-group} have concrete interpretations in a community health application.  Each community is a geographically distinct neighborhood served by a single health clinic.  For an individual $(x,a,c,y) \sim F_D$, the variable $x$ represents that individual's private health data, $a$ may represent a protected attribute such as race or gender, $c$ is the individual's local health clinic.  Finally $y$ represents whether or not the individual is ill and is therefore 
"qualified" to access medical resources to treat that illness.  

We are interested in selecting an \textbf{outcome predictor},
$\widehat{Y}: \mathcal{X} \times \mathcal{A} \times \mathcal{C} \rightarrow \mathcal{Y}$  for community group $D$ such that
for any individual $(x,a,c,y) \sim F_D$ we have $
\widehat{Y}(x,a,c) = y$ with a high probability.  In particular, let $(1-\Delta) \in (0,1)$ denote  a specified accuracy  level, then
the outcome predictor is $\Delta$-accurate if 
${\rm Pr}_D \left\{ \widehat{Y}(X,A,C) = Y \right\} \geq 1-\Delta$. 
With these definitions and notational conventions we can now formalize the specific notions of group fairness and community fairness.

\begin{definition}
\label{def:EO} The outcome predictor
$\widehat{Y} : \mathcal{X} \times \mathcal{A} \times \mathcal{C} \rightarrow \mathcal{Y}$ for community group $D$ satisfies
\textbf{equal opportunity} if and only if 
\begin{equation}\label{eq:EO}
\begin{split}
{\rm Pr}_D \left\{ \widehat{Y}(X,A,C) = 1\, | \, Y=1, A=1 \right\} = 
{\rm Pr}_D \left\{ \widehat{Y}(X,A,C) = 1 \, | \, Y=1, A=0 \right\}
\end{split}
\end{equation}
\end{definition}

Definition \ref{def:EO} asserts that the probability of
the outcome predictor correctly predicting a positive outcome for  an individual
$(x,a,c,y)$ from $D$ who qualifies for the positive outcome (i.e. $y=1$) is independent of the individual's protected attribute $a$.
\begin{remark}
\label{remark:local & global fairness}
   This paper defines group fairness in terms of Equal Opportunity over the \textit{global distribution}. It solves a different problem than \cite{chu2021fedfair}, which focuses fairness over each \textit{local distribution}. Global fairness and local fairness are not equivalent. See the Appendix \ref{App:local fairness vs global fairness} for more details.

\end{remark}

\begin{definition}
\label{def:CF} The outcome predictor 
$\widehat{Y} : \mathcal{X} \times \mathcal{A} \times \mathcal{C} \rightarrow \mathcal{Y}$ for community group $D$ satisfies
\textbf{community fairness} if and only if for any $j,k \in \mathcal{C}$, we have
\begin{equation}\label{eq:CF}
\begin{split}
{\rm Pr}_D \left\{ 
\widehat{Y}(X,A,C) = Y \, | \, C=j \right\} = 
{\rm Pr}_D \left\{
\widehat{Y}(X,A,C) = Y \, | \, C =k \right\}
\end{split}
\end{equation}
\end{definition}

Definition \ref{def:CF} asserts that the probability 
of the outcome predictor correctly predicting an individual's qualified outcome is independent of which community the individual belongs to.

This paper develops a post-processing FL algorithm that selects an outcome predictor
$\widetilde{Y} : \mathcal{X} \times \mathcal{A} \times \mathcal{C} \rightarrow \mathcal{Y}$ that satisfies both \emph{community fairness} and \emph{group fairness}.  If a $\Delta$-accurate predictor exists that achieves community and group fairness on community group $D$, then we say the community group is $\Delta$-equalizable.  This paper will also establish necessary conditions for a community group to be $\Delta$-equalizable.

\section{Achieving Group and Community Fairness }
\label{sec:lp}

Let $D = (X,A,C,Y)$ be a community group and 
consider the loss function $\ell : \mathcal{Y} \times \mathcal{Y} \rightarrow \{0,1\}$ that takes values 
\begin{eqnarray}
\ell(\widetilde{y},y) =
\mathbbm{1}(\widetilde{y} \neq y)
\label{eq:loss}
\end{eqnarray}
 for any $\widetilde{y},y \in \mathcal{Y}$, where $\mathbbm{1}(\cdot)$ is the indicator function.  A \textbf{fair outcome predictor} is any map
$\widetilde{Y} : \mathcal{X} \times \mathcal{A} \times \mathcal{C} \rightarrow \mathcal{Y}$ that satisfies the equal opportunity equation (\ref{eq:EO}) and community fairness equation (\ref{eq:CF}) with respect to community group $D$.
The following proposition asserts that if an outcome predictor
$\widetilde{Y}$ satisfies the following optimization problem in equation (\ref{eq:optimization-problem}), then $\widetilde{Y}$ must be a fair outcome predictor with respect to community group $D$.  


\begin{proposition}
\label{prop:1}
Consider the community group, $D = (X,A,C,Y)$, and the binary loss function, $\ell$, in equation (\ref{eq:loss}).  If the outcome predictor $\widetilde{Y} : \mathcal{X} \times \mathcal{A} \times \mathcal{C} \rightarrow  \mathcal{Y}$ satisfies the following optimization problem
\begin{eqnarray}
\begin{array}{ll}
\text{minimize} & \mathbb{E}_{D} \left[ \ell(\widetilde{Y}(X,A,C),Y) \right] \\
\text{with respect to} & \widetilde{Y}: \mathcal{X} \times \mathcal{A} \times \mathcal{C} \rightarrow \mathcal{Y} \\
\text{subject to} & {\rm Pr}_D (\widetilde{Y}(X,A,C) = 1 \, | \, Y=1, A=0 )  = {\rm Pr}_D (\widetilde{Y}(X,A,C)=1 \, | \, Y=1, A=1), \\
&  \text{${\rm Pr}_D(\widetilde{Y}(X,A,C)=Y \, | \, C = j)$} = \text{${\rm Pr}_D (\widetilde{Y}(X,A,C)=Y \, | \, C=k)$,}\quad  \forall j,k \in \mathcal{C}
\end{array}
\label{eq:optimization-problem}
\end{eqnarray}

then $\widetilde{Y}$ is a \textbf{fair outcome predictor}.
\end{proposition}

\vspace{0.1in}
\noindent{\bf Proof:} Any $\widetilde{Y}$ 
that solves optimization problem must satisfy the given constraints.  Since these constraints are equation (\ref{eq:EO}) and (\ref{eq:CF}) for equal opportunity and community fairness, respectively, the outcome predictor must also be fair with respect to community group $D$.
 $\diamondsuit$
 
 It is not yet clear if the optimization problem in  equation (\ref{eq:optimization-problem}) actually has a solution.  To obtain conditions for the existence of a fair outcome predictor, we will first show that equation (\ref{eq:optimization-problem}) can be recast as a linear program.  
 The existence of a fair outcome predictor is then equivalent to that linear program having non-negative solutions.  
 
 Let $\widehat{Y} : \mathcal{X} \times \mathcal{A} \times \mathcal{C} \rightarrow \mathcal{Y}$ be an \emph{optimal} outcome predictor that minimizes the expected value of the indicator loss function in equation (\ref{eq:loss}) with respect to community group $D$.  
 For notational convenience we will drop the arguments on the outcome predictors so we write $\widehat{Y}(X,A,C)$ (or $ \widetilde{Y}(X,A, C))$ as $\widehat{Y}$ (or $ \widetilde{Y}$).  
 For convenience we introduce the following notational conventions for the optimal outcome predictor's \textit{joint probability} of false or true positives and negatives:
\begin{equation}\label{eq:local_statistics_0}
    \begin{split}
    {\rm FN}^{ac} &={\rm Pr}_D \left\{ \widehat{Y} = 0, Y=1, A=a, C=c \right\} \\
      {{\rm TN}}^{ac} &= {\rm Pr}_D \left\{ \widehat{Y} = 0, Y=0, A=a, C=c \right\} 
     \\
     {\rm FP}^{ac} &= {\rm Pr}_D \left\{ \widehat{Y}=1, Y=0, A=a, C=c \right\} \\
    {\rm TP}^{ac} &= {\rm Pr}_D \left\{ \widehat{Y}=1, Y=1, A=a, C=c \right\} \\
    \end{split}
\end{equation}
 We will also find it convenient to define the following statistics for the group community, $D$.
 \begin{equation}\label{eq:local_statistics_1}
 \begin{split}
 p_c &= {\rm Pr}_D ( C=c) \\
 \alpha &= {\rm Pr}_D(Y=1 \, , \, A=0) \\
 \beta &=  {\rm Pr}_D(Y=1  \, , \, A=1)
\end{split}
\end{equation}
 
 $p_c$ is the probability of a random individual being in community $c$, $\alpha$ is the probability of a random individual  being qualified and non-sensitive and $\beta$ is the probability of a random individual being qualified and sensitive.  
 
 Finally, the variables we will use to characterize our fair outcome predictor, $\widetilde{Y}$, will be 
 \begin{eqnarray}\label{eq: variable}
 z^{ac}_j &=& 
 {\rm Pr}_D \left\{
 \widetilde{Y}=\widehat{Y} \, | \, \widehat{Y}=j, A=a, C=c \right\}
 \end{eqnarray}
 So $z^{ac}_j$ is the probability that the fair predictor's outcome, $\widetilde{Y}$,  equals that of the optimal predictor's outcome, $\widehat{Y}$,  for an individual, $(x,a,c,y)$, for which the optimal predictor's output is $j \in \mathcal{Y}$, the sensitive attribute is $a\in \mathcal{A}$ and the community label is $c\in \mathcal{C}$.   
 
\begin{proposition}
\label{prop:lp}
 (Appendix \ref{App:proof of 3.2}) Let $\mathbf{z} \in \mathbb{R}^{4K}$ satisfy the following linear program (LP)
\begin{eqnarray}\label{eq: lp}
\begin{array}{ll}
\mbox{minimize:} &\mathbf{c}^T \mathbf{z} \\
\mbox{with respect to:}& \mathbf{z}  \in \mathbb{R}^{4K} \\
\mbox{subject to:} & \mathbf{Az} = \mathbf{b} \\
& 0 \leq \mathbf{z} \leq 1
\end{array}
\end{eqnarray}
where
\begin{eqnarray*}
\mathbf{c}^T &=& \left[ \begin{array}{cccc}
\mathbf{c}_1^T & \mathbf{c}_2^T & \cdots & \mathbf{c}^T_K \end{array} \right] \\
\mathbf{z}^T &=& \left[ \begin{array}{cccc}
\mathbf{z}_1^T & \mathbf{z}_2^T & \cdots & \mathbf{z}^T_K \end{array} \right]\\
\end{eqnarray*}
\renewcommand{\arraystretch}{1.5} 
\[
\mathbf{A} = {\displaystyle \left[ \begin{array}{cccccc}
\mathbf{m}_1^T & \mathbf{m}_2^T & \mathbf{m}_3^T & \cdots &
\mathbf{m}^T_{K-1}&  \mathbf{m}^T_K \\
-\frac{K-1}{K} \mathbf{n}_1^T & \frac{1}{K} \mathbf{n}_2^T & \frac{1}{K} \mathbf{n}_3^T & \cdots &  \frac{1}{K} \mathbf{n}_{K-1}^T & \frac{1}{K} \mathbf{n}_K^T \\
\frac{1}{K} \mathbf{n}_1^T & - \frac{K-1}{K} \mathbf{n}_2^T & \mathbf{n}_3^T & \cdots & \frac{1}{K} \mathbf{n}_{K-1}^T & \frac{1}{K} \mathbf{n}_K^T \\
\frac{1}{K} \mathbf{n}_1^T & \frac{1}{K} \mathbf{n}_2^T &
-\frac{K-1}{K} \mathbf{n}_3^T & \cdots & 
\frac{1}{K} \mathbf{n}_{K-1}^T & \frac{1}{K} \mathbf{n}_K^T \\
\vdots & \vdots & \vdots & \ddots & \vdots & \vdots \\
\frac{1}{K} \mathbf{n}_1^T & \frac{1}{K} \mathbf{n}_2^T &
\frac{1}{K} \mathbf{n}_3^T & \cdots & - \frac{K-1}{K} \mathbf{n}^T_{K-1} & \frac{1}{K} \mathbf{n}^T_K \\
\frac{1}{K} \mathbf{n}_1^T & \frac{1}{K} \mathbf{n}_2^T &
\frac{1}{K} \mathbf{n}_3^T & \cdots &  \frac{1}{K} \mathbf{n}^T_{K-1} & -\frac{K-1}{K} \mathbf{n}^T_K
\end{array} \right]}
\]
\begin{eqnarray*}
\mathbf{b}^T &=&\left[ \begin{array}{ccccc}
  {\displaystyle 
 \sum_{c=1}^K  \left( \frac{{\rm FN}^{1c}}{\beta} - \frac{{\rm FN}^{0c}}{\alpha} \right)} & 
 {\displaystyle \frac{1}{K}\sum_{c=1}^K (b_1 - b_c)} & 
 {\displaystyle \frac{1}{K} \sum_{c=1}^K (b_2-b_c)}  & 
 \cdots & 
{\displaystyle  \frac{1}{K} \sum_{c=1}^K (b_{K}- b_c) }
\end{array} \right]
\end{eqnarray*}
with 
\begin{eqnarray*}
\mathbf{c}_i^T &=& \left[ \begin{array}{cccc}
({\rm FN}^{0i}-{\rm TN}^{0i})& ({\rm FP}^{0i}-{\rm TP}^{0i})& ({\rm FN}^{1i}-{\rm TN}^{1i})&({\rm FP}^{1i}-{\rm TP}^{1i}) \end{array}\right]  \\
\mathbf{n}_i^T &=&\frac{1}{p_i} \left[ \begin{array}{cccc}
({\rm FN}^{0i}-{\rm TN}^{0i})& ({\rm FP}^{0i}-{\rm TP}^{0i})& ({\rm FN}^{1i}-{\rm TN}^{1i})&({\rm FP}^{1i}-{\rm TP}^{1i}) \end{array}\right]  \\
\mathbf{z}_i^T &=& \left[ \begin{array}{cccc}
 z_0^{0i}& z_1^{0i}& z_{0}^{1i}& z_1^{1i}\end{array}\right] \\
\mathbf{m}_i^T &=& \left[ \begin{array}{cccc}
    \frac{-{\rm FN}^{0i}}{\alpha} &  \frac{{\rm TP}^{0i}}{\alpha}&\frac{{\rm FN}^{1i}}{\beta}
     & \frac{-{\rm TP}^{1i}}{\beta}
\end{array}\right] \\
b_i &=&  \frac{1}{p_i} ({\rm TN}^{0i}+{\rm TP}^{0i}+{\rm TN} ^{1i}+{\rm TP}^{1i})
\end{eqnarray*}

for $i=1,2,\cdots K$. 

Let the solution of the linear program $\mathbf{z}$ be:
\begin{eqnarray*}
\mathbf{z}^T &=& \left[ \begin{array}{cccc}
\mathbf{z}_1^T & \mathbf{z}_2^T & \cdots & \mathbf{z}^T_K \end{array} \right]
\end{eqnarray*}
with 
\begin{eqnarray*}
    \mathbf{z}_i^T &=& \left[ \begin{array}{cccc}
 z_0^{0i}& z_1^{0i}& z_{0}^{1i}& z_1^{1i}\end{array}\right] 
\end{eqnarray*}

Then the outcome predictor
$\widetilde{Y}_{\widehat{Y}, \mathbf{z}} : \mathcal{X} \times \mathcal{A} \times \mathcal{C} \rightarrow \mathcal{Y}$ taking values
\begin{equation}\label{fair_outcome}
\begin{split}
\quad \quad &  \widetilde{Y}_{\widehat{Y}, \mathbf{z}}(x,a,c)= \begin{cases} 0 & \text{with probability } z_0^{ac} \\ 1 & \text{with probability } 1-z_0^{ac}\end{cases} 
\quad\text{if  }  \widehat{Y}(x,a,c)=1 \nonumber \\
\text{or}\\
 \quad \quad & \widetilde{Y}_{\widehat{Y}, \mathbf{z}}(x,a,c)= \begin{cases} 1 & \text{with probability } z_1^{ac} \\ 0 & \text{with probability } 1-z_1^{ac}\end{cases}
\quad \text{if  } \widehat{Y}(x,a,c)=0\nonumber\\ 
\end{split}
\end{equation}
is a fair outcome predictor.
\end{proposition}
\vspace{0.1in}
 \begin{remark}
 \label{remark: extension to multi-class problem}
    Our post-FFL algorithm can be extended to multi-class classification tasks, i.e., \(\mathcal{Y} = \{1, 2, \cdots, N\}\). The linear program for multi-class classification is detailed in Appendix \ref{App: multi-class extension}.
\end{remark}
 \begin{remark}\label{relaxed_fairness}
    In general, we may not necessarily need to achieve strictly identical performance across all communities. It is worth noting that the optimization shown in (\ref{eq: lp}) can incorporate a more relaxed version of equal opportunity and community fairness constraints.
    {\small
    \begin{equation}
    \begin{split}
       |{\rm Pr_D}(\Tilde{Y}=1|Y=1,A=0)-{\rm Pr_D}(\Tilde{Y}=1|Y=1,A=1)|&\leq \epsilon\\
       \forall k\in \mathcal{C}, |{\rm Pr}(\widetilde{Y}\neq Y|C=k)-\frac{1}{K} \sum_{c=1}^{K}  {\rm Pr}(\widetilde{Y}\neq Y|C=c)| &\leq \delta
    \end{split}
    \end{equation}
    }
  In such cases, the linear program with the relaxed fairness constraints is:
  \begin{eqnarray}\label{eq: lp_relax}
    \begin{array}{ll}
    \mbox{minimize:} &\mathbf{c}^T \mathbf{z} \\
    \mbox{with respect to:}& \mathbf{z}  \in \mathbb{R}^{4K} \\
    \mbox{subject to:} & \mathbf{b}-\boldsymbol{\epsilon} \leq \mathbf{Az} \leq \mathbf{b}+\boldsymbol{\epsilon} \\
    & 0 \leq \mathbf{z} \leq 1
    \end{array}
    \end{eqnarray}
 where, \begin{eqnarray*}
\boldsymbol{\epsilon}^T &=& \left[ \begin{array}{cccc}
\epsilon & \delta & \cdots & \delta \end{array} \right]
 \end{eqnarray*}
   By solving the linear program with different $\boldsymbol{\epsilon}\in \mathbbm{R}^{(K+1)}$, we can precisely control the degree of community fairness and group fairness, setting $\boldsymbol{\epsilon}=\mathbf{0}$ gives us the outcome predictor that strictly satisfies equal opportunity and community fairness.$\diamondsuit$
\end{remark}
 We can write the linear program (\ref{eq: lp}) in the \textit{standard form} by introducing a set of slack variables $\mathbf{s}\in \mathbb{R}^{4K}$:
 \begin{eqnarray*}
\mathbf{s}^T &=& \left[ \begin{array}{cccc}
\mathbf{s}_1^T & \mathbf{s}_2^T & \cdots & \mathbf{s}^T_K \end{array} \right]    
 \end{eqnarray*}
 with
 \begin{eqnarray*}
     \mathbf{s}_i^T &=& \left[ \begin{array}{cccc}
 s_0^{0i}& s_1^{0i}& s_{0}^{1i}& s_1^{1i}\end{array}\right] \\ 
 \end{eqnarray*}
The variables we need to solve for in linear program (\ref{eq: lp}) represent probabilities, thus, $0 \leq z_j^{ac} \leq 1$. It is equivalent to: 
\begin{equation}\label{eq:slack_variables}
    \begin{split}
        z_j^{ac}+s_j^{ac}&=1\\
        z_j^{ac}, s_j^{ac} &\geq 0
    \end{split}
\end{equation}

Combine the linear program (\ref{eq: lp}) with (\ref{eq:slack_variables}), the standard form of the linear program (\ref{eq: lp}) is:
\vspace{-0.05in}
\begin{eqnarray}\label{eq: standard_lp}
\begin{array}{ll}
\mbox{minimize:} &\mathbf{\Bar{c}}^T \mathbf{\Bar{z}} \\
\mbox{with respect to:}& \mathbf{\Bar{z}}  \in \mathbb{R}^{8K} \\
\mbox{subject to:} & \mathbf{\Bar{A} \Bar{z}} = \mathbf{\Bar{b}} \\
&  \mathbf{\Bar{z} } \geq 0
\end{array}
\end{eqnarray}
with
\begin{eqnarray*}
\mathbf{\Bar{c}} ^T &=& \left[ \begin{array}{cccc}
\mathbf{c}^T & \mathbf{0}  \end{array} \right] \in \mathbb{R}^{8K }\\
\mathbf{\Bar{z}}^T &=& \left[ \begin{array}{cccc}
\mathbf{z}^T & \mathbf{s}^T  \end{array} \right] \in \mathbb{R}^{8K}\\
\mathbf{\Bar{A}} &=& \left[
\begin{array}{cc}
  \mathbf{A}   & \mathbf{0}  \\
    \mathbf{I} & \mathbf{I}
\end{array}
\right] \in \mathbb{R}^{(5K+1)\times 8K}\\
\mathbf{\Bar{b}} ^T &=& \left[ \begin{array}{cccc}
\mathbf{b}^T & \mathbf{1}_{4K}^ T\end{array} \right] \in \mathbb{R}^{5K+1}
\end{eqnarray*}

where $\mathbf{1}_{4K}\in \mathbb{R}^{4K}$ is all $1$ vector.
\begin{theorem}\label{theo: existence}
(Appendix \ref{App:proof_theo_3.4})
    The linear program (\ref{eq: standard_lp}) always has non-negative solutions.

\end{theorem}
Theorem \ref{theo: existence} indicates that the linear program (\ref{eq: lp}), which is equivalent to (\ref{eq: standard_lp}), always has a solution. Thus, there always exists fair outcome predictors satisfying both equal opportunity and community fairness. We next demonstrate the necessary conditions for the existence of a $\Delta$-accurate fair outcome predictor. An outcome predictor: $\widetilde{Y}: \mathcal{X}\times \mathcal{A}\times \mathcal{C}\rightarrow \mathcal{Y} $ is a $\Delta$-accurate fair outcome predictor if:
{\small
\begin{equation} \label{eq: epsilon_fair_predictor}
    \begin{split}
        {\rm Pr}_D (\widetilde{Y}\neq Y)&\leq \Delta\\
        {\rm Pr}_D (\widetilde{Y}=1| Y=1, A=0)&= {\rm Pr}_D (\widetilde{Y}=1|Y=1, A=1)\\
        \forall k\in \mathcal {C}, {\rm Pr}_D (\widetilde{Y}\neq Y|C=k)&=\frac{1}{K} \sum_{c=1}^{K}  {\rm Pr}_D (\widetilde{Y} \neq Y|C=c)
    \end{split}
\end{equation}
}

\begin{theorem} \label{theo:fair_outcome_predictor}
(Appendix \ref{App: proof_theo_3.5})  
The three conditions in (\ref{eq: epsilon_fair_predictor}) cannot hold simultaneously if  $\Delta < -\|\mathbf{\Bar{c}}\|_\infty \frac {\|\mathbf{\Bar{A}}^T\mathbf{\Bar{b}}\|_2}{\underline{\sigma}^2}+\sum_{c=1}^{K}({\rm TN}^{0c}+ {\rm TP}^{0c}+ {\rm TN}^{1c}+ {\rm TP}^{1c}) $ , where, $\underline{\sigma}$ is the smallest singular value of the matrix $\mathbf{\Bar{A}}$. 
\end{theorem}

 Theorem \ref{theo:fair_outcome_predictor} establishes a necessary condition for the existence of an $\Delta$-accurate fair outcome predictor, based on the statistics of the data represented by  $\mathbf{\Bar{A}},\mathbf{\Bar{b}},\mathbf{\Bar{c}}$. The following theorem estimates the accuracy lost in a $\boldsymbol{\epsilon}$-relaxed fair predictor.

\begin{theorem} (Appendix \ref{App: Proof_of_theo_3.6})\label{Theorem: accuracy_loss}
    Let $\mathbf{z}\in \mathbbm{R}^{4K}$ be the solution of the linear program (\ref{eq: lp_relax}) with a predefined $\boldsymbol{\epsilon}$, then the minimum accuracy we  lose for improving both community fairness and group fairness under post-FFL is $\mathbf{c}^T(\mathbf{z}-\mathbf{1}_{4k})$, where, $\mathbf{1}_{4K}$ is all $1$ vector.
\end{theorem}
\vspace{-0.1in}
Theorem \ref{Theorem: accuracy_loss} above can serve as a tool for evaluating the accuracy we lose for improving group fairness and community fairness
\vspace{-0.1in}

\section{Post-FFL: Fair Outcome Predictor in FL}
This section shows how the linear program (\ref{eq: lp}) can be used within a federated learning framework to construct a fair outcome predictor. An overview of post-FFL is shown in Fig. \ref{Fig:framework}. The black arrows illustrate the exchange of local model's parameters and the global model's parameters during \textit{FedAvg} training. The blue arrows depict the post-processing workflow following \textit{FedAvg}, where local communities send their local statistics to the global server. The global server constructs a linear program and sends the solution back to the local communities. Each local community uses the decision tree, as shown on the right of Fig.\ref{Fig:framework}, to make fair outcome predictions. Below, we provide a list itemizing the concrete training steps of post-FFL: 
\begin{figure*}[ht]
\begin{center}
\centerline{\includegraphics[width=7.1in]{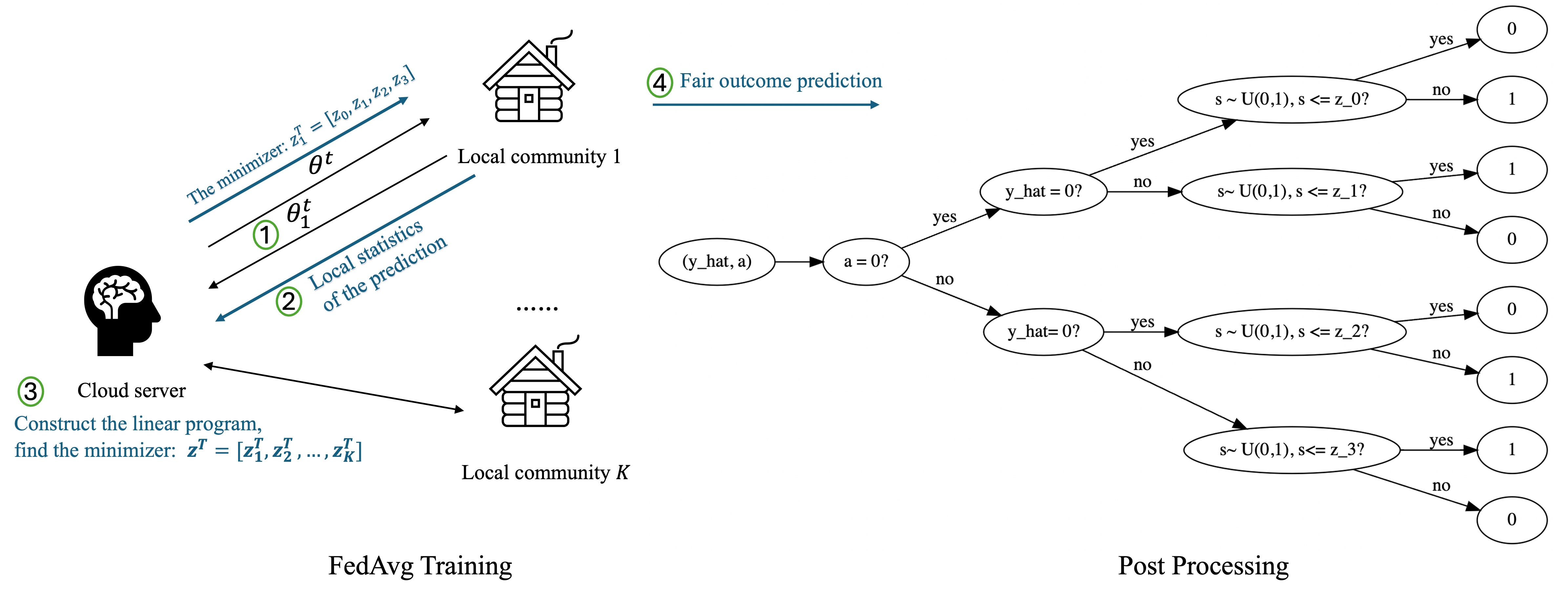}}
\caption{Overview of the proposed post-FFL framework.}
\label{Fig:framework}
\end{center}
\end{figure*}
{\small
\begin{algorithm*}[t]
    \caption{Fair Outcome Predictor}
    \label{alg:1}

        \textbf{Input:} The optimal outcome predictor: $\widehat{Y}: \mathcal{X}\times \mathcal{A} \times \mathcal{C}\rightarrow \mathcal{Y}$, the community $k$'s corresponding minimizer:
        
        \quad \quad $\mathbf{z}_k^T = \left[ \begin{array}{cccc}
         z_0^{0k}& z_1^{0k}& z_{0}^{1k}& z_1^{1k}\end{array}\right] $
         
        \textbf{Output:}  Fair outcome predictor $\widetilde{Y}_{\widehat{Y},\mathbf{z}_k}: \mathcal{X}\times \mathcal{A} \times \mathcal{C}\rightarrow \mathcal{Y}$     


        \quad 1. randomly sample $s \sim U(0,1)$, the uniform distribution between (0,1)

        \quad 2. Construct $\widetilde{Y}_{\widehat{Y},\mathbf{z}_k}(x,a,k)$ as

        \quad \quad $\widetilde{Y}_{\widehat{Y},\mathbf{z}_k}(x,a,k)=\left\{\begin{array}{l}
        0,  \text { if } (a=0) \text{ and } ((\widehat{Y}=0 \text{ and } s\leq z_0^{0k}) \text { or } (\widehat{Y}=1 \text { and } s> z_1^{0k})) \\
        1,  \text { if }(a=0) \text{ and } ((\widehat{Y}=0 \text{ and } s> z_0^{0k}) \text { or } ((\widehat{Y}=1 \text { and } s\leq z_1^{0k})) \\
        0,  \text { if } (a=1) \text{ and }((\widehat{Y}=0 \text{ and } s\leq z_0^{1k}) \text { or } (\widehat{Y}=1 \text { and } s> z_1^{1k}))  \\
        1,  \text { if } (a=1) \text{ and } ((\widehat{Y}=0 \text{ and } s> z_0^{1k}) \text { or } (\widehat{Y}=1 \text { and } s \leq z_1^{1k}))
        \end{array}\right.$

        \textbf{return} $\widetilde{Y}_{\widehat{Y},\mathbf{z}_k}$

\end{algorithm*}
}


\begin{enumerate}
    \item \textbf{Training an Optimal Outcome Predictor using \textit{FedAvg}}: The server and communities collaboratively train an optimal predictor $\widehat{Y}: \mathcal{X}\times \mathcal{A}\times \mathcal{C} \rightarrow \mathcal{Y}$ using the \textit{FedAvg} algorithm \cite{mcmahan2017communication}. The participating community trains a local model at time $t$ and sends the local model's parameters, $\boldsymbol{\theta}_c^t \in \mathbb{R}^n$, to the global server. The global server then aggregates the local models into a global model parameter, $\boldsymbol{\theta}^t = \sum_{c=1}^{K} p_c \boldsymbol{\theta}_c^t$, and sends it back to the local communities. This iterative process continues until the parameters, $\boldsymbol{\theta}$, of the optimal predictor $\widehat{Y}$ are determined. The aggregation weight, $p_c = \Pr_D(C=c)$, is estimated from the dataset as shown in (\ref{eq:est_p_c}).

    \begin{equation}\label{eq:est_p_c}
        \begin{split}
            p_c= \frac{\text{number of samples in community $c$}}{\text{number of total samples}}
        \end{split}
    \end{equation}
    \item \textbf{Local Prediction and Probability Calculation}: Each local community generates predictions $\widehat{Y}(X,A,C)\in \mathcal{Y}$, computes local statistics ${\rm Pr}_D({\widehat{Y}=y, Y=y', A=a \mid C=c})$ for all $(y, y', a) \in \{0,1\}^3$ as specified in (\ref{eq: est_statistics}), and then transmits these probabilities to the global server.
    {\small
    \begin{equation} \label{eq: est_statistics}
        \begin{split}
            &{\rm Pr}_D (\hat{Y}=y, Y=y', A=a \mid C=c)= \frac{\text{number of samples with ($\hat{Y}=y, Y=y', A=a$) in community $c$}}{\text{number of samples in community c}}
        \end{split}
    \end{equation}
    }
    \item \textbf{Constructing and Solving the Linear Program}: The global server computes the parameters defined in Equations (\ref{eq:local_statistics_0}) and (\ref{eq:local_statistics_1}) using the probabilities sent by the communities.
    The parameters ${\rm FN^{ac}}, {\rm TN^{ac}}, {\rm FP^{ac}}, {\rm TP^{ac}}$ in (\ref{eq:local_statistics_0}) are computed based on:
    \begin{equation*}
        \begin{split}
            &{\rm Pr}_D \left\{ \widehat{Y} = y, Y=y', A=a, C=c \right\} = p_c {\rm Pr}_D \left\{ \widehat{Y} = y, Y=y', A=a \mid C=c \right\}
        \end{split}
    \end{equation*}
    The parameters $\alpha$ and $\beta$ in (\ref{eq:local_statistics_1}) are computed as:
    \begin{eqnarray*}
        \alpha &=& {\rm Pr}_D(Y=1, A=0) = \sum_{c=1}^{K}({\rm FN}^{0c} + {\rm TP}^{0c})\\
        \beta &=&  {\rm Pr}_D(Y=1, A=1) = \sum_{c=1}^{K}({\rm FN}^{1c} + {\rm TP}^{1c})
    \end{eqnarray*}
    Using the above parameters, the global server constructs the linear program (\ref{eq: lp}), finds the minimizer $\mathbf{z}$:
    \begin{equation*}
        \mathbf{z}^T = \left[ \begin{array}{cccc}
        \mathbf{z}_1^T & \mathbf{z}_2^T & \cdots & \mathbf{z}_K^T \end{array} \right]\\
    \end{equation*}
    and then sends the corresponding minimizer $\mathbf{z}_k^T$ to community $k$, where $k = 1, 2, \cdots, K$.
    \item \textbf{Fair Outcome Predictor}: The local community $k$, $(k=1,2,\cdots K)$ employs Algorithm~\ref{alg:1} to make fair predictions. The received minimizer $\mathbf{z}_k^T$ indicates the probability that the fair predictor's outcome  equals the optimal predictor's outcome for community $k$. 
\end{enumerate}

Algorithm \ref{alg:1} provides a community dependent randomized function that is used to decide whether to accept or deny the prediction from the optimal model. The output of the optimal model, combined with the randomized function, will yield a fair outcome predictor.

\section{Experiments}
We conduct experiments on the real-world dataset to demonstrate the effectiveness of the post-FFL. Section 5.1 verifies that the proposed post-FFL can simultaneously enforce both notions of fairness for binary and multi-class classification tasks. It also shows that our framework can flexibly adjust the degree of fairness. Section 5.2 compares our framework with other existing fair federated algorithms. Our framework outperforms current in-processing techniques in terms of achieving both notions of fairness, communication efficiency, and computational cost. Section 5.3 presents results under different data heterogeneity settings. Section 5.4 explores the effects of the local epoch and batch size in \textit{FedAvg} algorithm.

\paragraph{Dataset.} We use three real-world datasets: the Adult, the Diabetes, and the HM10000 to demonstrate the effectiveness of post-FFL.

The \textbf{Adult} dataset \cite{asuncion2007uci} consists of 6 numerical features (age, final weight, education number, etc.) and 8 categorical features (work class, education, gender, race, etc.) and is used to predict a \textit{binary label}. The label indicates whether an individual earns more than 50K/year. We set gender as the sensitive attribute. Following the federated setting in \cite{li2020federated, mohri2019agnostic}, we split the dataset into \textit{two communities}: one is the PhD community, in which all individuals are PhDs, and the other is the non-PhD community.

The \textbf{Diabetes} dataset \cite{strack2014impact} contains 10 numerical features (time in hospital, number of procedures, etc.), 40 binary features (race, gender, age range, admission source, diabetMed, etc.), and is used to predict a \textit{binary label}. The label indicates whether a patient will be readmitted within 30 days. We set the group ``older (aged over 60) African-American females'' as the sensitive group. We split the data into \textit{seven communities} based on their admission source. Communities 1, 2, 3, 4, 5, 6, and 7 represent samples admitted from the Emergency Room, Physician Referral, NULL, Transfer from a hospital, Transfer from another healthcare facility, Clinic Referral, Transfer from a Skilled Nursing Facility, and Others, respectively.

The \textbf{HM10000} dataset \cite{Tschandl2018} comprises dermatoscopic images from different populations and is used to predict \textit{seven classes}. The seven skin diagnostic classes are: akiec (Actinic Keratoses and Intraepithelial Carcinoma), bcc (Basal Cell Carcinoma), bkl (Benign Keratosis), df (Dermatofibroma), nv (Melanocytic Nevi), mel (Melanoma), and vasc (Vascular Skin Lesions). We set individuals aged over 60 as the protected group. Equal Opportunity in this multi-class problem is defined as the true positive rate of Basal Cell Carcinoma being equal across the protected and unprotected groups. We divide the dataset into \textit{five communities}. These communities are defined as follows: Community 1 represents individuals aged under 40; Community 2, those aged 40 to 50; Community 3, individuals aged 50 to 60; Community 4, those aged 60 to 70; and Community 5 includes those older than 70 and individuals of unknown ages.

\paragraph{Evaluation Metrics.}
We evaluate the model's performance from the following perspectives:
\begin{itemize}
    \item  \textbf{Model utility:} We use the model's \textbf{Avg-Acc} to measure its utility.
    The Avg-Acc is the weighted average accuracy across all communities.
    
    \item \textbf{Group fairness:} Group fairness, as defined in Definition \ref{def:EO}, requires that the true positive rates are the same for sensitive and non-sensitive groups. We measure group fairness using the \textbf{Equal Opportunity Difference (EOD)}, which is defined as the difference in true positive rates between sensitive and non-sensitive groups. 
    \item \textbf{Community fairness:} Community fairness, as defined in Definition \ref{def:CF}, requires that the model has similar accuracy across all communities. We measure community fairness using  \textbf{Accuracy Disparity (AD)} which is the difference between the highest and lowest accuracy across all communities.
\end{itemize}

The full experiments details (including model and hyperparameters) are in Appendix \ref{App: exp_details}
\subsection{Fairness of post-FFL}

In this section, we verify that the proposed framework can simultaneously enforce group fairness and community fairness within a FL platform. We demonstrate our framework can control the degree of group fairness and community fairness by adjusting the \(\epsilon\) and \(\delta\) in the linear program (\ref{eq: lp_relax}) respectively. Smaller \(\epsilon\) and \(\delta\) values will lead to a fairer outcome predictor. Setting \((\epsilon=0, \delta=0)\) results in a predictor that strictly achieves both group fairness and community fairness. We further show that Theorem (\ref{Theorem: accuracy_loss}) within our framework allows one to evaluate the accuracy loss incurred while improving group fairness and community fairness.

\paragraph{Achieving group fairness and community fairness.}
Table \ref{table:Fedavg and post_FFL} reports 
 Avg-Acc, EOD, and AD of \textit{FedAvg} \cite{mcmahan2017communication} and post-FFL $(\epsilon = 0, \delta = 0)$ for all three datasets.  These results show post-FFL reduced EOD and AD by 66.0\%-87.5 \% over \textit{FedAvg}, thereby simultaneously achieving high levels of group and community fairness.

\begin{table}[ht]
\caption{The EOD, AD and Avg-Acc of \textit{FedAvg}  and our post-FFL. The 95\% confidence interval (mean$\pm$ margin of error) of all measurements is computed over 5 runs.
 }
\begin{center}
{
\begin{tabular}{c|c|c|c|c}
\hline
{Dataset} & {Frameworks} & {EOD} & {AD} & {Avg-Acc} \\
\hline
\multirow{2}{*}{Adult} & Fedavg & 0.10 $\pm$ 0.02 & 0.14 $\pm$ 0.11 & 0.85 $\pm$ 0.00 \\
\cline{2-5}
                       & Post-FFL & \textbf{0.02 $\pm$ 0.00\tiny($\downarrow${80.00\%})} & \textbf{0.03 $\pm$ 0.01\tiny($\downarrow${78.57\%})} & 0.75 $\pm$ 0.03 \\
\hline
\multirow{2}{*}{Diabetes} & Fedavg & 0.07$\pm$ 0.03 & 0.09 $\pm$ 0.02 & 0.83 $\pm$ 0.01 \\
\cline{2-5}
                          & Post-FFL & \textbf{0.02$\pm$0.02\tiny($\downarrow${74.43\%})} & \textbf{0.03 $\pm$ 0.01\tiny($\downarrow${66.67\%})} & 0.82$\pm$ 0.01 \\
\hline
\multirow{2}{*}{HM1000} & Fedavg & 0.08 $\pm$ 0.02 & 0.33 $\pm$ 0.03 & 0.74 $\pm$ 0.02 \\
\cline{2-5}
                          & Post-FFL & \textbf{0.01$\pm$0.02\tiny($\downarrow${87.50\%})} & \textbf{0.08 $\pm$ 0.01\tiny($\downarrow${75.75\%})} & 0.58$\pm$ 0.04 \\
\hline
\end{tabular}
}
\end{center}
\label{table:Fedavg and post_FFL}
\end{table}

\paragraph{Flexibility of Adjusting the Trade-off Between Fairness and Accuracy.}
Table \ref{table:post_FFL_with_different_parameter} reports the
 Avg-Acc, EOD, and AD for different settings of \((\epsilon, \delta)\) on all three datasets.  These results show that with fixed $\epsilon$, decreasing $\delta$ reduces AD and Avg-Acc, indicating the model is fairer with respect to community fairness but has poorer performance. Similarly, with a fixed \(\delta\), the degree of equal opportunity was effectively controlled by adjusting \(\epsilon\). 
 These results support our assertion that post-FFL is effective tool for trading off fairness and model accuracy.

\begin{table*}[h]
\centering
\caption{post-FFL performance for a range of $(\epsilon,\delta)$ relaxed fair predictors.}
\begin{small}
\begin{tabular}{c|c|c|c|c||c|c|c}

\hline
{Dataset}&$(\epsilon, \delta)$ & EOD & AD & Avg-Acc& empirical accuracy loss & estimated accuracy loss& estimated error\\  \hline
\multirow{8}{*}{Adult}&(0.00, 0.00) & 0.016 & 0.012 & 0.780 &0.074&0.074&0.000 \\ 
&(0.00, 0.02) & 0.008 & 0.053 & 0.804 &0.050&0.056&0.006\\
&(0.00, 0.04) & 0.001 & 0.091& 0.841&0.013&0.012&0.001 \\ 
&(0.02, 0.00) & 0.033 & 0.014 & 0.766& 0.088&0.095&0.007\\ 
&(0.02, 0.02) & 0.030 & 0.051 & 0.802&0.051&0.056&0.004 \\ 
&(0.02, 0.04) & 0.017 & 0.090 & 0.837&0.017&0.018&0.001\\ 
&(0.04, 0.00) & 0.049& 0.016 & 0.758& 0.095&0.096&0.001 \\ 
&(0.04, 0.02) & 0.029 & 0.056 & 0.797&0.057&0.057&0.000 \\ 
&(0.04, 0.04) & 0.044 & 0.094& 0.839&0.014&0.018&0.004 \\ \hline
\multirow{8}{*}{Diabetes}&(0.00, 0.00) & 0.008 & 0.021 & 0.812 &0.022&0.022&0.000 \\ 
&(0.00, 0.02) & 0.034 & 0.067 & 0.830 &0.002&0.001&0.001\\ 
&(0.00, 0.04) & 0.047 & 0.080& 0.831&0.003&0.003&0.000 \\ 
&(0.02, 0.00) & 0.040 & 0.013 & 0.815& 0.018&0.022&0.004\\ 
&(0.02, 0.02) & 0.025 & 0.048 & 0.830&0.004&0.001&0.003 \\ 
&(0.02, 0.04) & 0.027 & 0.077 & 0.832&0.002&0.000&0.002\\ 
&(0.04, 0.00) & 0.057& 0.025 & 0.815& 0.018&0.022&0.004 \\ 
&(0.04, 0.02) & 0.006 & 0.070 & 0.831&0.002&0.001&0.001 \\ 
&(0.04, 0.04) & 0.032 & 0.076& 0.831&0.002&0.00&0.002 \\ \hline
\multirow{8}{*}{HM10000}&(0.00, 0.00) & 0.012 & 0.089 & 0.623 &0.146&0.131&0.015 \\ 
&(0.00, 0.05) & 0.010 & 0.176 & 0.687&0.083&0.077&0.006\\ 
&(0.00, 0.10) & 0.034 &0.242 & 0.736&0.032&0.023&0.009 \\ 
&(0.02, 0.00) & 0.060 & 0.060 & 0.617& 0.152&0.131&0.021\\ 
&(0.02, 0.05) & 0.043 & 0.160 & 0.682&0.087&0.077&0.010 \\ 
&(0.02, 0.10) & 0.044 & 0.249 & 0.723&0.045&0.023&0.022\\ 
&(0.04, 0.00) & 0.060& 0.045 & 0.616& 0.152&0.131&0.021 \\ 
&(0.04, 0.05) & 0.050 & 0.154 & 0.669&0.100&0.077&0.023 \\ 
&(0.04, 0.10) & 0.057 & 0.242& 0.728&0.041&0.023&0.017 \\ \hline
\end{tabular}
\end{small}
\label{table:post_FFL_with_different_parameter}
\end{table*}
\paragraph{Estimating the accuracy trade-off for fairness.}
The last three columns of
Table \ref{table:post_FFL_with_different_parameter} compare a relaxed post-FFL model's accuracy against the theoretical estimate of accuracy loss in   Theorem (\ref{Theorem: accuracy_loss}).  The columns show a theoretical estimate that was within (74.65$\pm$11.34)\% of the empirical accuracy for the particular $(\epsilon,\delta)$ relaxations over all three datasets.  
This suggests Theorem 
(\ref{Theorem: accuracy_loss}) can estimate the accuracy lost when enforcing group and community fairness.
\subsection{Comparison with other algorithms} \label{sec:compare with other algorithms}

This subsection compares post-FFL with other existing fair federated learning algorithms. We did not find prior work that tries to simultaneously achieve group fairness and community fairness in a federated setting. The most relevant prior work employs in-processing (rather than post-processing) techniques to achieve either equal opportunity or fair resource allocation (community fairness) in a federated setting. We used \textit{q-FedAvg} \cite{li2019fair} and \textit{FairFed} \cite{ezzeldin2023fairfed} as  baselines for community fairness and group fairness, respectively. 

 A full description of all baselines is provided in Appendix \ref{app: baselines}
\paragraph{Achieving Both Notions of Fairness.} Table \ref{table:compare_with_alternatives} examines the performance of post-FFL on  the \textit{Adult} and \textit{HM10000} datasets using three evaluation metrics (EOD, AD, Avg-Acc). These results show that 
 \textit{FairFed} improves group fairness (EOD) by $83.96$-$98.91$\% over \textit{FedAvg} on the two datasets while making community fairness (AD) worse.  
 The table results show that 
\textit{q-FedAvg} improves community fairness (AD) by $14.15$-$98.39$\% over \textit{FedAvg} on two datasets with little change in group fairness (EOD).  
The table shows that post-FFL was able to significantly improve group (EOD) and community (AD) fairness over both datasets to levels consistent with
\textit{FairFed} (EOD) and \textit{q-FedAvg} (AD).

\begin{table}[ht]
\centering
\begin{small}
\caption{ EOD, AD and Avg-Acc of all algorithms. }
\label{table:compare_with_alternatives}
\begin{tabular}{c|c|c|c|c|c|c}
\hline
&\multicolumn{3}{c|}{Adult}& \multicolumn{3}{c}{HM1000}\\ \hline
Algorithms & EOD & AD & Avg-Acc& EOD & AD & Avg-Acc\\ \hline
Initial \textit{FedAvg}& 0.106& 0.124 &{0.854}&0.092&0.311&{0.769}\\ \hline
    Our \textit{post-FFL} $(\epsilon=0, \delta=0)$& {0.016 \small($\downarrow${84.91\%})}& 0.012\small($\downarrow${90.32\%})& 0.780&{0.012\small($\downarrow${86.96\%})}&{0.089\small($\downarrow${71.38\%})}&0.623 \\ \hline
\textit{q-FedAvg}& 0.337\small($\uparrow${217.92\%}) & 0.002\small($\downarrow${98.39\%}) & 
 0.811&0.071\small($\downarrow${22.82\%})&0.267\small($\downarrow${14.15\%})&0.730 \\ \hline
\textit{FairFed}& 0.017\small($\downarrow${83.96\%}) & 0.362\small($\uparrow${191.94\%}) & 
 0.846 &{0.001\small($\downarrow${98.91\%})}& 0.420\small($\uparrow${35.05\%})&0.673 \\ \hline
\end{tabular}
\end{small}
\vspace{-0.1in}
\end{table}


\paragraph{Communication Efficiency and Computational Cost}
Table \ref{table:one_round_time} reports
the number of communication rounds, computation time for one round, and total computation time for 
\textit{FedAvg}, \textit{FairFed} and \textit{q-FedAvg} for the Adult and HM10000 datasets.  Note that because
post-FFL simply post-processes the models obtained by \textit{FedAvg}, the communication and computation cost of post-FFL is accurately represented by \textit{FedAvg}'s cost.  So the results
in Table \ref{table:one_round_time} are actually comparing post-FFL's computational cost to the inprocessing fair FL algorithms \textit{FairFed} and \textit{q-FedAvg}.  The computation time results were obtained on a 16-Core 4.00 GHz AMD RYZEN Threadripper Pro 5955WX Processor. The average time is calculated over 30 communication rounds for each method. 

The results in Table \ref{table:one_round_time} show that the number of communication rounds required by post-FFL is 2-3 orders of magnitude lower than that required by \textit{q-FedAvg} and one order of magnitude lower than what \textit{FairFed} required.
The post-FFL's computation time for a single round was comparable to that used by \textit{q-FedAvg}.  When compared to
\textit{FairFed}, post-FFL single round time was comparable on the more complex HM10000 dataset and an order of magnitude faster on the Adult dataset. The table shows for the total amount of time needed for algorithm convergence, post-FFL was 2 order of magnitude faster on the Adult data set and about one order faster on the HM10000 dataset for both in-processing fair FL algorithms.
Training curves for a specific run  in Fig.\ref{fig:convergence} of Appendix \ref{app: baselines} provide a more "graphic" illustration of post-FFL's convergence speed.
These results, therefore, support our assertion that post-FFL (FedAvg) has significantly lower computational and communication cost than prior in-processing approaches to fair FL.

\begin{table}[ht]
\centering
\caption{Number of communication rounds for convergence, average time taken for completing one round of update and total time for convergence of \textit{FedAvg}, \textit{q-FedAvg} and\textit{FairFed}.}
{\small
\begin{tabular}{c|c|c|c|c|c|c}
\hline
& \multicolumn{3}{c|}{Adult}& \multicolumn{3}{c}{HM10000}\\
\hline
Objectives & \textit{FedAvg} & \textit{q-Fedavg} & \textit{FairFed} & \textit{FedAvg} & \textit{q-Fedavg} & \textit{FairFed}\\ \hline
\# of convergence rounds &$\approx$ \textbf{5}& $> 1000$&
$\approx${20}& $\approx$ \textbf{10}&$>200$& $\approx 15$ \\ \hline
Avg-time for 1 round &\textbf{0.631s}&0.639s&16.428s&\textbf{10.619s}&10.863s&14.016s\\ 
\hline
Total time for convergence &\textbf{$\approx$3.155s}& $>$639s &$\approx$328.56s&\textbf{$\approx$106.19s}&$>$2172.6s& $\approx$210.24s\\
\hline
\end{tabular}
}
\vspace{-0.22in}
\label{table:one_round_time}
\end{table}

\subsection{Experimental Result With Different Data Heterogeneity Level}

We present results with various degrees of data heterogeneity across all communities. We introduce community heterogeneity by adjusting the mixture of individuals with specific target and sensitive attributes. The results for \textit{Adult} dataset with two communities are presented, and the results for HM10000 are in Appendix \ref{app:additional results on HM1000}.

\begin{figure*}[h]
\caption{The class proportion of the three scenarios.}
\label{Fig: different heterogenity level}
\subfigure[Scenario 1]{
\begin{minipage}[t]{0.33\linewidth}
\centering
\includegraphics[width=2.4in]{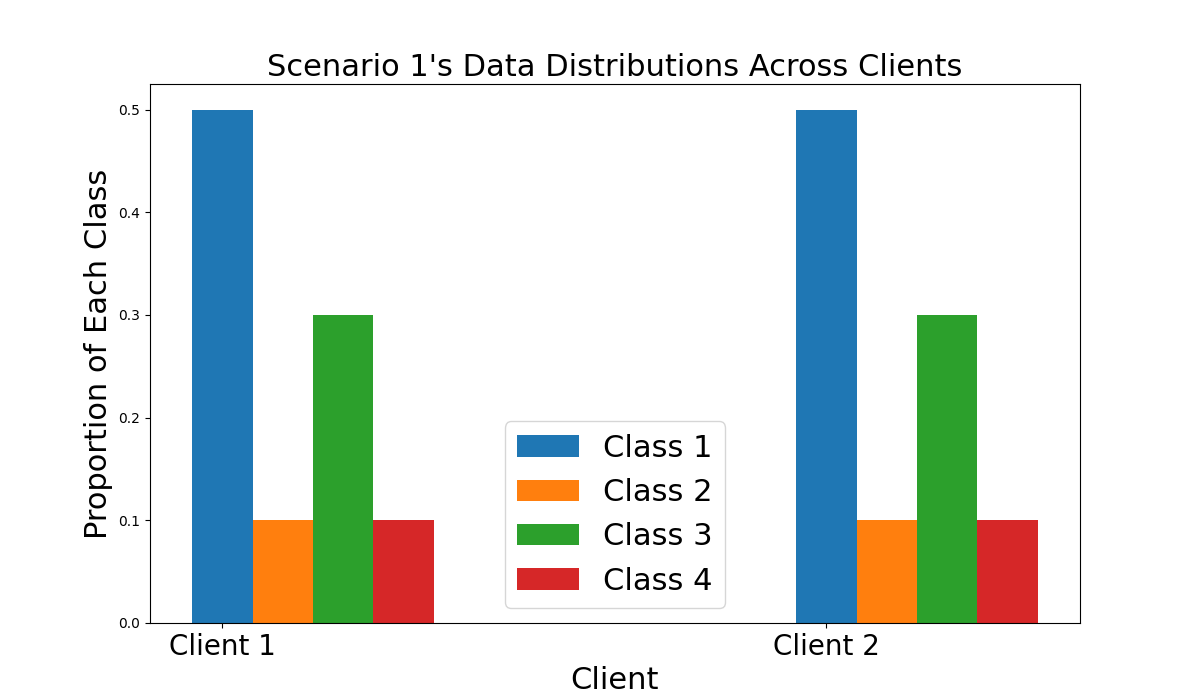}
\end{minipage}%
}%
\subfigure[Scenario 2]{
\begin{minipage}[t]{0.33\linewidth}
\centering
\includegraphics[width=2.4in]{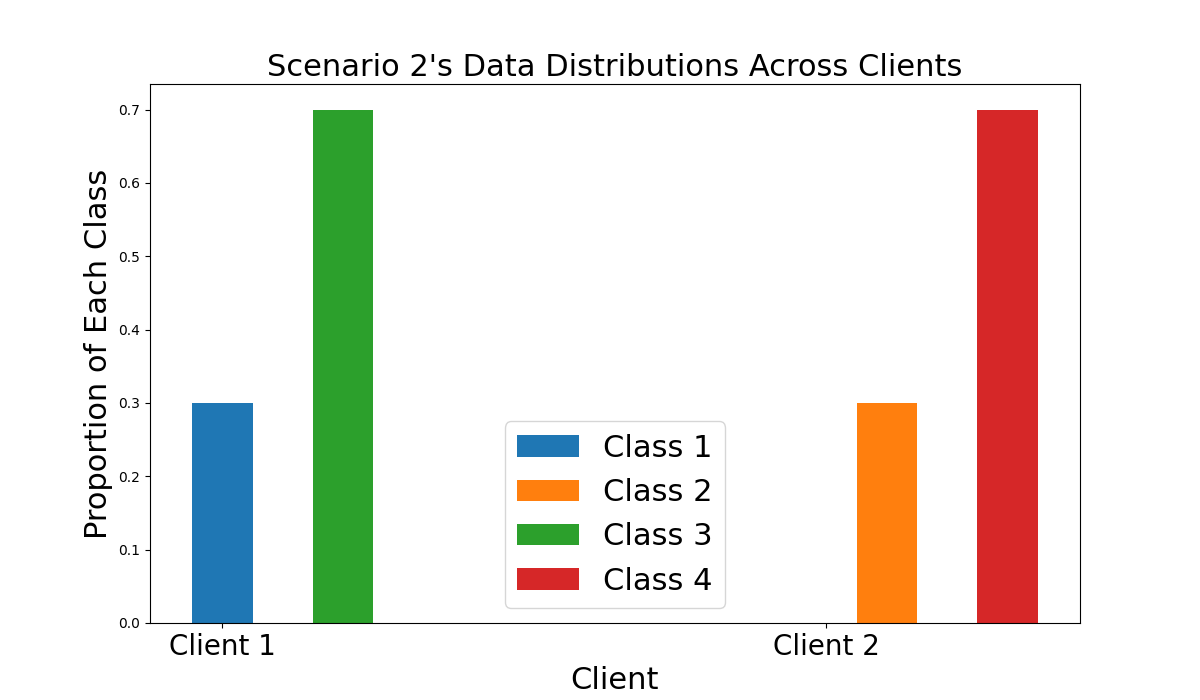}
\end{minipage}%
}%
\subfigure[Scenario 3]{
\begin{minipage}[t]{0.33\linewidth}
\centering
\includegraphics[width=2.4in]{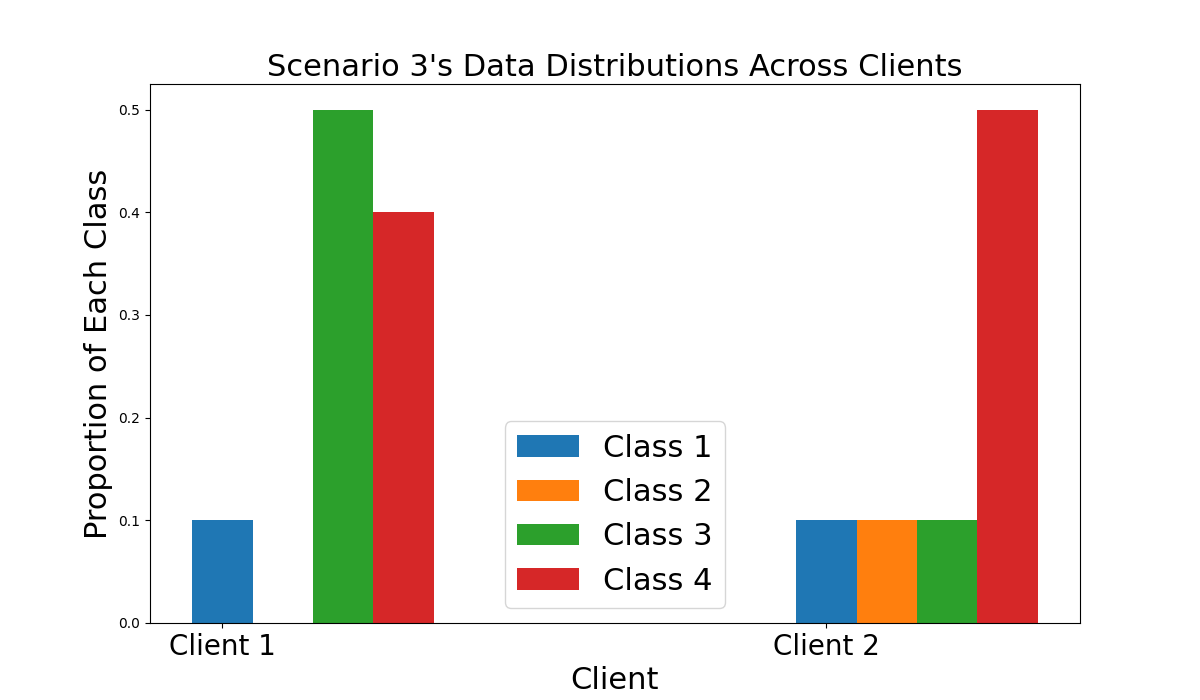}
\end{minipage}%
}%
\end{figure*}

We define class 1 as positive male $(y = 1, s = 1)$, class 2 as positive female $(y = 1, s = 0)$, class 3 as negative male $(y = 0, s = 1)$, and class 4 as negative female $(y = 0, s = 0)$. We sample the local communities' datasets so that the class proportion of each community is as shown in Fig. \ref{Fig: different heterogenity level}. For scenario 1, data is i.i.d. distributed across all communities.
The sensitive attribute is totally skewed across two communities in scenario 2. Both the target and sensitive attributes are imbalanced across communities in scenario 3.

 Table \ref{tab: different_i_i_d_setting} presents the measurements for all scenarios. We observe resents these results which show that post-FFL reduced EOD 
and AD by 50\%-100 \% over FedAvg across all scenarios, thereby simultaneously achieving high levels of group and
community fairness across different heterogeneity scenarios. Compared to Scenario 1, Scenario 2 has a higher degree of data heterogeneity, resulting in a greater initial AD for \textit{FedAvg}. Consequently, the accuracy traded for fairness in Scenario 2 (0.110) is greater than in Scenario 1 (0.108). This demonstrates that as the level of data heterogeneity increases, the accuracy traded to ensure fairness also increases.

\begin{table}[ht]
    \centering
    \caption{Experimental results with different degrees of heterogeneity (Adult)}
    \label{Table: data hetergeneity 2}
    \resizebox{\textwidth}{!}{
    \begin{tabular}{c|ccccc|ccccc|c}
        \hline
        \multirow{2}{*}{Level of heterogeneity } & \multicolumn{5}{c|}{FedAvg} & \multicolumn{5}{c|}{Post-FFL} & \multirow{2}{*}{\begin{tabular}{c} 
        Accuracy trade for fairness 
        \end{tabular}} \\
        \cline{2-6} \cline{7-11}
        & Avg-Acc & EOD & Acc @1 & Acc@2 & AD & Avg-Acc & EOD & Acc @1 & Accc @2 & AD & \\
        \hline
        Scenario 1 & 0.817 & 0.080 & 0.826 & 0.808 & 0.017 &0.799 & 0.013 & 0.809& 0.788 & 0.021 & 0.018 \\
        Scenario 2 & 0.855 & 0.320 & 0.784 & 0.929 & 0.145 & 0.745 & 0.046 & 0.745& 0.751 & 0.006 & 0.110 \\
        Scenario 3 & 0.875 & 0.242 & 0.840 & 0.910 & 0.070 & 0.837 & 0.036 & 0.846 & 0.828 & 0.018 & 0.038 \\
        \hline
    \end{tabular}
    }
    \label{tab: different_i_i_d_setting}
\end{table}
\subsection{Experimental results with various local Batch Sizes and Epochs.}
Since post-FFL post-processes models obtained using FedAvg, it is important to assess the sensitivity
of post-FFL’s fairness with respect to FedAvg’s hyperparameters, in particular the batch size \(B\) and
number of local epochs \(E\) used to obtain the local models. 
\(B\) represents the fraction of data samples used for each local model update. \(B = 1\) indicates that the full local dataset is treated as a single minibatch. \(E\) is the number of training epochs in each communication round. (\(B = 1, E = 1\)) corresponds exactly to FedSGD \cite{mcmahan2017communication}. We present the results for the \textit{ Adult} dataset. The class proportion of communities used corresponds to Scenario 2 of Table \ref{tab: different_i_i_d_setting}. 

The results with different \(B,E\) are in Table \ref{tab:different_B_E}. We observe that EOD is reduced to $(0.01\pm0.01)$ and AD are reduced to $(0.01\pm0.01)$ after applying post-FFL for all different settings. Post-FFL can simultaneously ensure group fairness and community fairness, regardless of the local epochs \(E\) and local batch size \(B\). Post-FFL's has low sensitivity to batch size and number of local epochs


\begin{table}[h]
    \centering
    \caption{Experimental results with different $\mathrm{B}$ and $\mathrm{E}$ in FedAvg algorithm (Adult)}
    \resizebox{\textwidth}{!}{
    \begin{tabular}{c|ccccc|ccccc|c}
        \hline
        \multirow{2}{*}{Settings} & \multicolumn{5}{c|}{FedAvg} & \multicolumn{5}{c|}{Post-FFL} & \multirow{2}{*}{Accuracy trade for fairness} \\
        \cline{2-6} \cline{7-11}
        & Avg-Acc & EOD & Acc @1 & Acc@2 & AD & Avg-Acc & EOD & Acc @1 & Accc @2 & AD & \\
        \hline
        $\mathrm{B}=1, \mathrm{E}=1$ (FedSGD) & 0.841 & 0.350 & 0.719 & 0.924 & 0.205 & 0.67 & 0.019 & 0.679 & 0.661 & 0.018 & 0.171 \\
        $\mathrm{B}=1, \mathrm{E}=5$ & 0.860 & 0.241 & 0.768 & 0.921 & 0.153 & 0.722 & 0.012 & 0.714 & 0.729 & 0.015 & 0.138 \\
        $\mathrm{B}=1 / 3, \mathrm{E}=1$ & 0.863 & 0.440 & 0.799 & 0.923 & 0.124 & 0.668 & 0.003 & 0.669 & 0.668 & 0.001 & 0.195 \\
        $\mathrm{B}=1 / 3, \mathrm{E}=5$ & 0.855 & 0.300 & 0.801 & 0.912 & 0.11 & 0.759 & 0.008 & 0.759 & 0.745 & 0.014 & 0.096 \\
        $\mathrm{B}=1 / 6, \mathrm{E}=1$ & 0.862 & 0.493 & 0.799 & 0.924 & 0.125 & 0.679 & 0.001 & 0.670 & 0.687 & 0.017 & 0.183 \\
        $\mathrm{B}=1 / 6, \mathrm{E}=5$ & 0.868 & 0.229 & 0.796 & 0.895 & 0.098 & 0.724 & 0.015 & 0.724 & 0.724 & 0.000 & 0.144 \\
        \hline
    \end{tabular}%
    }
    \label{tab:different_B_E}
\end{table}

\section{Conclusions and Limitations}
\label{conclusion and limitation}
In this work, we propose the post-FFL framework, which simultaneously achieves group and community fairness in FL. Experiments on real-world datasets demonstrate that post-FFL allows users to adjust the degree of fairness, and the theoretical results associated with it can predict the accuracy cost for fairness. Post-FFL outperforms existing baselines in fair federated learning in terms of fairness improvement, communication efficiency, and computational cost.

Post-FFL selects a random outcome predictor from a fixed set of classifiers. The fixed nature of this set may limit the achievable accuracy. Existing literature \cite{hardt2016equality, xian2023fair} derives fair predictor from a pre-trained score function may lead to a more optimal model. We will explore the score-based approach in FL in future work.

\section*{Acknowledgments}
The authors
acknowledge the partial financial support of the Lucy Family Institute for Data and Science and the National Science Foundation - grant
CNS-2228092.


\newpage

\bibliography{reference}
\bibliographystyle{icml2024}

\newpage
\appendix
\onecolumn
\section{Global Fairness and Local Fairness are not Equivalent.}
\label{App:local fairness vs global fairness}
As stated in Remark \ref{remark:local & global fairness}, ensuring fairness in each \textit{local distribution} does not guarantee global fairness, as local fairness and global fairness are not equivalent.

The local fairness with respect to Equal Opportunity (EO) over each \textit{local distribution} in FL is defined as:
\begin{eqnarray}
\label{eq:local_EO}
\forall c\in \mathcal{C}, {\rm Pr}(\widehat{Y}=1|Y=1,A=0,C=c)={\rm Pr}(\widehat{Y}=1|Y=1,A=1,C=c)
\end{eqnarray}

Our paper studies group fairness over the \textit{global distribution}, i.e.,
\begin{eqnarray}
\label{eq:global_EO}
{\rm Pr}(\widehat{Y}=1|Y=1,A=0)={\rm Pr}(\widehat{Y}=1|Y=1,A=1)
\end{eqnarray}

It is obvious that if the model is fair over the global distribution, it is not necessarily fair over the local distributions, i.e., equation (\ref{eq:global_EO}) does not imply equation (\ref{eq:local_EO}). Conversely, we cannot ensure global fairness by simply training locally fair models, i.e., equation (\ref{eq:local_EO}) does not imply equation (\ref{eq:global_EO}).
The reason is as follows: 

Assume we have a model, $\widehat{Y}: \mathcal{X}\times \mathcal{A}\times \mathcal{C} \rightarrow \mathcal{Y}$, that is fair within each community, i.e., satisfying equation (\ref{eq:local_EO}).
The global Equal Opportunity Difference (EOD) for the model $\widehat{Y}$ is:
\begin{eqnarray*}
{\rm EOD}:&=& {\rm Pr}(\widehat{Y}=1|Y=1,A=0)-{\rm Pr}(\widehat{Y}=1|Y=1,A=1)\\
&=& \sum_{c\in \mathcal{C}} \left[{\rm Pr}(\widehat{Y}=1,C=c|Y=1,A=0)-{\rm Pr}(\widehat{Y}=1,C=c|Y=1,A=1)\right]\\
&=&\sum_{c\in \mathcal{C}}\left[{\rm Pr}(\widehat{Y}=1|Y=1,A=0,C=c)\cdot {\rm Pr}(C=c|Y=1,A=0) \right.\\
&& \quad  \quad \left. -{\rm Pr}(\widehat{Y}=1|Y=1,A=1,C=c)\cdot {\rm Pr}(C=c|Y=1,A=1) \right]
\end{eqnarray*}
\noindent Given that the model, $\widehat{Y}$ satisfies EO in (\ref{eq:local_EO}), we can substitute ${\rm Pr}(\widehat{Y}=1|Y=1,A=1,C=c)$ with ${\rm Pr}(\widehat{Y}=1|Y=1,A=0,C=c)$:
{\small
\begin{eqnarray*}
{\rm EOD}&=& \sum_{c\in \mathcal{C}}\left[\left( {\rm Pr}(C=c|Y=1,A=0)-{\rm Pr}(C=c|Y=1,A=1)\right)  \cdot {\rm Pr}(\widehat{Y}=1|Y=1,A=0,C=c) \right]
\end{eqnarray*}
}

The global EOD is not necessarily zero, i.e., the model is not fair over the \textit{global distribution}. The global EOD is zero when the community label $C$ and the sensitive attribute $A$ are conditionally independent, given $Y=1$. However, in most real-world scenarios of interest to us, the sensitive attribute, for example, "race" and the community label indicates which geographic "neighborhood" the household is located in are highly correlated due to a history of racial segregation. 
  So we cannot, in general, achieve global fairness simply by enforcing local fairness at the client level.

\section{The LP for Multi-Class Classification.}\label{App: multi-class extension}
As mentioned in Remark \ref{remark: extension to multi-class problem}, our approach can be extended to multi-class problems, where $x \in \mathcal{X}$, $a \in \mathcal{A} = \{0,1\}$, $c \in \mathcal{C} = \{1, 2, \ldots, K\}$, and $y \in \mathcal{Y} = \{0, 1, \ldots, N\}$.
 
In multi-class problems, the variables we use to characterize the fair outcome predictor $\Tilde{Y}$ are,
 \begin{eqnarray*}
 z^{ac}_{kj} &=& 
 {\rm Pr} \left\{
 \widetilde{Y}=k, | \, \widehat{Y}=j, A=a, C=c \right\},  \forall k\in \mathcal{Y}, j\in \mathcal{Y}, a\in \mathcal{A}, c\in \mathcal{C}
 \end{eqnarray*}
We set $\alpha={\rm Pr}(Y=1,A=0)$, $\beta= {\rm Pr}(Y=1,A=1), p_c= \rm{Pr}(C=c)$ and use 
 the the joint statistics over $\hat{Y}, Y, A, C$: $p_{kj}^{ac}={\rm Pr}(Y=k, \hat{Y}=j, A=a, C=c)$ to construct the linear program.

The LP of multi-class problem is:
\begin{eqnarray*}
\begin{array}{ll}
\mbox{maximize:} & \sum\limits_{c\in \mathcal{C}} \sum\limits_{a \in \mathcal{A}}\sum\limits_{j\in \mathcal{Y}}\sum\limits_{k\in \mathcal{Y}}p_{kj}^{ac}z^{ac}_{kj} \\[0.5cm]
\mbox{with respect to:}& z_{kj}^{ac}\in \mathbb{R}, \forall k\in \mathcal{Y}, j\in \mathcal{Y}, a\in \mathcal{A}, c\in \mathcal{C},\\[0.3cm]
\mbox{subject to:} & \sum\limits_{c\in \mathcal{C}}\sum\limits_{j \in \mathcal{Y}}\frac{p_{1j}^{0c}}{\alpha}\cdot z_{1j}^{0c}-\sum\limits_{c \in \mathcal{C}} \sum\limits_{j \in \mathcal{Y}}\frac{p_{1j}^{1c}}{\beta}z_{1j}^{1c}=0\\[0.5cm]
& \forall c \in \mathcal{C}, \sum\limits_{a \in \mathcal{A}} \sum\limits_{j \in \mathcal{Y}} \sum\limits_{k \in \mathcal{Y}} \frac{p_{kj}^{ac}}{p_c} \cdot z_{kj}^{ac} - \frac{1}{K} \sum\limits_{c \in \mathcal{C}} \sum\limits_{a \in \mathcal{A}} \sum\limits_{j \in \mathcal{Y}} \sum\limits_{k \in \mathcal{Y}} \frac{p_{kj}^{ac}}{p_c} \cdot z_{kj}^{ac}=0 \\[0.5cm]
& \forall c\in \mathcal{C}, a \in \mathcal{A}, j\in \mathcal{Y}, \sum\limits_{k \in \mathcal{Y}} z_{kj}^{ac}=1\\ [0.5cm]
& \forall k\in \mathcal{Y}, j\in \mathcal{Y}, a\in \mathcal{A}, c\in \mathcal{C}, 0 \leq z_{kj}^{ac} \leq 1

\end{array}
\end{eqnarray*}

Then, the outcome predictor: $\widetilde{Y}: \mathcal{X} \times \mathcal{A} \times \mathcal{C} \rightarrow \mathcal{Y}$   that takes values:  
\begin{equation}\label{eq:multi-class_fair_outcome_predictor}
\begin{split}
&\text{If }  \widehat{Y}(x,a,c)=j, A=a, C=c: \widetilde{Y}(x,a,c)= k 
\quad \text{with probability}\quad z^{ac}_{kj}
\end{split}
\end{equation}
is a fair outcome predictor.

\noindent{\bf Proof:}
 We first show that the outcome predictor in (\ref{eq:multi-class_fair_outcome_predictor}) satisfies Equal Opportunity:

 ${\rm} Pr(\Tilde{Y}=1|Y=1, A=a)$ can be expressed as:
 \begin{equation}
     \begin{split}
         {\rm Pr}(\widetilde{Y}=1|Y=1,A=a)&= \sum_{c\in \mathcal{C}}\frac{{\rm Pr}(\widetilde{Y}=1,Y=1,C=c,A=a)}{{\rm Pr}(Y=1,A=a)}\\
         &= \sum_{c \in \mathcal{C}}\sum_{j\in \mathcal{Y}} \frac{{\rm Pr(\widehat{Y}}=1,Y=1, C=c, A=a)}{{\rm Pr}(Y=1,A=a)}\cdot {\rm Pr}(\widetilde{Y}=1|\widehat{Y}=j, A=a, C=c)\\
         &= \sum_{c\in \mathcal{C}} \sum_{j \in \mathcal{Y}} \frac{p_{1j}^{ac}}{{\rm Pr}(Y=1,A=a)}z_{1j}^{ac}
     \end{split}
 \end{equation}
 Thus, the first equation of the LP is:
\begin{equation}
    \begin{split}
        0&=\sum_{c\in \mathcal{C}} \sum_{j \in \mathcal{Y}} \frac{p_{1j}^{0c}}{\alpha}z_{1j}^{0c}-\sum_{c\in \mathcal{C}} \sum_{j \in \mathcal{Y}} \frac{p_{1j}^{1c}}{\beta}z_{1j}^{1c}\\
        &= {\rm Pr} (\widetilde{Y}=1|Y=1,A=0)-{\rm Pr} (\widetilde{Y}=1|Y=1,A=1)
    \end{split}
\end{equation}
which indicates Equal Opportunity.

We then show that the fair outcome predictor satisfies Community Fairness:

The accuracy of the predictor of community $c$ is ${\rm Pr}(\widetilde{Y}=Y|C=c)$ and can be written as:
\begin{equation}
    \begin{split}
        {\rm Pr}(\widetilde{Y}=Y|C=c)&= \sum_{a\in \mathcal{A}} \sum_{k \in \mathcal{Y}} {\rm Pr}(\widetilde{Y}=k, Y=k, A=a|C=c)\\
        &= \sum_{a \in \mathcal{A}} \sum_{k \in \mathcal{Y}}\frac{{\rm Pr}(\widetilde{Y}=k, Y=k, A=a, C=c)}{{\rm Pr}(C=c)}\\
        & = \sum_{a\in \mathcal{A}}\sum_{k\in \mathcal{Y}}\sum_{j\in \mathcal{Y}}\frac{{\rm Pr}(\widehat{Y}=k, Y=k, A=a, C=c)}{{\rm Pr}(C=c)}\cdot {\rm Pr}(\widetilde{Y}=k|\widehat{Y}=j, A=a, C=c)\\
        &=\sum_{a\in \mathcal{A}}\sum_{k\in \mathcal{Y}}\sum_{j\in \mathcal{Y}}\frac{p_{kj}^{ac}}{p_c}\cdot z_{kj}^{ac}
    \end{split}
\end{equation}
Thus, the next equation of the LP is:
\begin{equation}
    \begin{split}
        \forall c \in \mathcal{C}, 0&=\sum_{a\in \mathcal{A}}\sum_{k\in \mathcal{Y}}\sum_{j\in \mathcal{Y}}\frac{p_{kj}^{ac}}{p_c}\cdot z_{kj}^{ac}-\frac{1}{K}\sum_{c\in \mathcal{C}}\sum_{a\in \mathcal{A}}\sum_{k\in \mathcal{Y}}\sum_{j\in \mathcal{Y}}\frac{p_{kj}^{ac}}{p_c}\cdot z_{kj}^{ac}\\
        &= {\rm Pr}(\widetilde{Y}=Y|C=c)-\frac{1}{K}\sum_{c\in \mathcal{C}} {\rm Pr}(\widetilde{Y}=Y|C=c) 
    \end{split}
\end{equation}
which indicates community fairness.

The objective function we maximize is:
\begin{equation}
    \begin{split}
        \sum\limits_{c\in \mathcal{C}} \sum\limits_{a \in \mathcal{A}}\sum\limits_{j\in \mathcal{Y}}\sum\limits_{k\in \mathcal{Y}}p_{kj}^{ac}z^{ac}_{kj}
        &= \sum \limits_{c \in \mathcal{C}}{\rm Pr}(\widetilde{Y}=Y|C=c)\cdot p_c\\
        &= \sum \limits_{c \in \mathcal{C}}{\rm Pr}(\widetilde{Y}=Y|C=c)\cdot {\rm Pr}(C=c)\\
        &= {\rm Pr}(\widetilde{Y}=Y)
    \end{split}
\end{equation}
which is the accuracy over the global distribution.

Since the variable $z^{ac}_{kj}$ represents probability, we have:
\begin{equation}
    \begin{split}
        & \forall c\in \mathcal{C}, a \in \mathcal{A}, j\in \mathcal{Y}, \sum\limits_{k \in \mathcal{Y}} z_{kj}^{ac}=1\\
        & \forall k\in \mathcal{Y}, j\in \mathcal{Y}, a\in \mathcal{A}, c\in \mathcal{C}, 0 \leq z_{kj}^{ac} \leq 1
    \end{split}
\end{equation}
which are the last two equations of the constraints.

Thus, the outcome predictor is a fair outcome predictor. $\diamondsuit$

\section{Theoretical Proofs} 
\subsection{Proof of Proposition \ref{prop:lp}}\label{App:proof of 3.2}
We first show that the outcome predictor $\widetilde{Y}_{\widehat{Y}, \mathbf{z}}: \mathcal{X}\times \mathcal{A}\times \mathcal{C}\rightarrow \mathcal{Y}$ satisfies equal opportunity:

The probability ${\rm Pr}_D (\widetilde{Y}_{\widehat{Y}, \mathbf{z}}=1|Y=1, A=a)$ can be extended as:
\begin{equation}\label{eq:extension}
\begin{split}
&{\rm Pr}_D(\widetilde{Y}_{\widehat{Y}, \mathbf{z}}=1|Y=1,A=a) \\
=& \frac{{\rm Pr}_D(\widetilde{Y}_{\widehat{Y}, \mathbf{z}}=1, Y=1, A=a)}{{\rm Pr}_D (Y=1, A=a)} \\
=& \frac{ \sum_{c=1}^{K} {\rm Pr}_D (\widetilde{Y}_{\widehat{Y}, \mathbf{z}}=1,Y=1, A=a, C=c)}{{\rm Pr}_D (Y=1, A=a)}
\\
=& \frac{\sum_{c=1}^{K}({\rm Pr}_D(\widehat{Y}=1, Y=1, A=a, C=c)\cdot {\rm Pr}_D (\widetilde{Y}_{\widehat{Y}, \mathbf{z}}=\widehat{Y}|\widehat{Y}=1, A=a, C=c))}{{\rm Pr}_D(Y=1,A=a)}\\
+& \frac{\sum_{c=1}^{K}({\rm Pr}_D(\widehat{Y}=0, Y=1, A=a, C=c)\cdot {\rm Pr}_D (\widetilde{Y}_{\widehat{Y}, \mathbf{z}}\neq \widehat{Y}|\widehat{Y}=0, A=a, C=c))}{{\rm Pr}_D(Y=1,A=a)}\\
=& \frac{\sum_{c=1}^{K}({\rm Pr}_D(\widehat{Y}=1, Y=1, A=a, C=c)\cdot z_1^{ac}}{{\rm Pr}_D(Y=1,A=a)}
+ \frac{\sum_{c=1}^{K}({\rm Pr}_D(\widehat{Y}=0, Y=1, A=a, C=c)\cdot (1-z_0^{ac})}{{\rm Pr}_D(Y=1,A=a)}\\
=& \frac{\sum_{c=1}^{K}{\rm TP}^{ac}\cdot z_1^{ac}}{{\rm Pr}_D(Y=1,A=a)}
+ \frac{\sum_{c=1}^{K}{\rm FN}^{ac}\cdot (1-z_0^{ac})}{{\rm Pr}_D(Y=1,A=a)}\\
\end{split}
\end{equation}

We can now calculate the \textit{Equal opportunity Difference} of the outcome predictor $\widetilde{Y}_{\widehat{Y}, \mathbf{z}}$, which is defined as: 
${\rm Pr}_D(\widetilde{Y}_{\widehat{Y}, \mathbf{z}}=1|Y=1,A=0)- {\rm Pr}_D(\widetilde{Y}_{\widehat{Y}, \mathbf{z}}=1|Y=1,A=1)$:
\begin{equation}\label{eq:equal_opportunity_difference}
\begin{split}
&{\rm Pr}_D(\widetilde{Y}_{\widehat{Y}, \mathbf{z}}=1|Y=1,A=0)- {\rm Pr}_D(\widetilde{Y}_{\widehat{Y}, \mathbf{z}}=1|Y=1,A=1)\\
=& \frac{\sum_{c=1}^{K}{\rm TP}^{0c}\cdot z_1^{0c}}{{\rm Pr}_D(Y=1,A=0)}
+ \frac{\sum_{c=1}^{K}{\rm FN}^{0c}\cdot (1-z_0^{0c})}{{\rm Pr}_D(Y=1,A=0)}-(\frac{\sum_{c=1}^{K}{\rm TP}^{1c}\cdot z_1^{1c}}{{\rm Pr}_D(Y=1,A=1)}
+ \frac{\sum_{c=1}^{K}{\rm FN}^{1c}\cdot (1-z_0^{1c})}{{\rm Pr}_D(Y=1,A=1)})\\
=& \frac{-\sum_{c=1}^{K}{\rm FN}^{0c}\cdot z_0^{0c}}{\alpha}+\frac{\sum_{c=1}^{K}{\rm TP}^{0c}\cdot z_1^{0c}}{\alpha}+ \frac{\sum_{c=1}^{K}{\rm FN}^{1c}\cdot z_0^{1c}}{\beta}-\frac{\sum_{c=0}^{K}{\rm TP}^{1c}\cdot z_1^{1c}}{\beta}+\frac{\sum_{c=1}^{K}{\rm FN}^{0c}}{\alpha}-\frac{\sum_{c=1}^{K}{\rm FN}^{1c}}{\beta}
\end{split}
\end{equation}

The first linear equation of $\mathbf{Az}=\mathbf{b}$ in (\ref{eq: lp}) is:
\begin{equation}\label{eq:first_lq}
 \begin{split}
     0=&\sum_{c=1}^{K} \mathbf{m}_c^T \mathbf{z}_c-\sum_{k=1}^K  \left( \frac{{\rm FN}^{1k}}{\beta} - \frac{{\rm FN}^{0k}}{\alpha} \right)\\
     =&\sum_{c=0}^{K} \left[ \begin{array}{cccc}
    \frac{-{\rm FN}^{0c}}{\alpha} &  \frac{{\rm TP}^{0c}}{\alpha}&\frac{{\rm FN}^{1c}}{\beta}
     & \frac{-{\rm TP}^{1c}}{\beta}
    \end{array}\right]^T\cdot \left[ \begin{array}{cccc}
    z_0^{0c}\\z_1^{0c}\\z_0^{1c}\\z_1^{1c}
    \end{array}\right]-\sum_{c=1}^K  \left( \frac{{\rm FN}^{1c}}{\beta} - \frac{{\rm FN}^{0c}}{\alpha} \right)\\
   =  & \frac{-\sum_{c=1}^{K}{\rm FN}^{0c}\cdot z_0^{0c}}{\alpha}+\frac{\sum_{c=1}^{K}{\rm TP}^{0c}\cdot z_1^{0c}}{\alpha}+ \frac{\sum_{c=1}^{K}{\rm FN}^{1c}\cdot z_0^{1c}}{\beta}-\frac{\sum_{c=0}^{K}{\rm TP}^{1c}\cdot z_1^{1c}}{\beta}+\frac{\sum_{c=1}^{K}{\rm FN}^{0c}}{\alpha}-\frac{\sum_{c=1}^{K}{\rm FN}^{1c}}{\beta}\\
&\text{Combine the above with the \textit{Equal Opportunity Difference}'s expression  (\ref{eq:equal_opportunity_difference}) }:\\
 &{\rm Pr}_D(\widetilde{Y}_{\widehat{Y}, \mathbf{z}}=1|Y=1,A=0)- {\rm Pr}_D(\widetilde{Y}_{\widehat{Y}, \mathbf{z}}=1|Y=1,A=1)=0
 \end{split}   
\end{equation}
We can see from (\ref{eq:first_lq}) the first linear equation leads a  outcome predictor that satisfies equal opportunity.

Then, we show that the outcome predictor $\widetilde{Y}_{\widehat{Y}, \mathbf{z}}: \mathcal{X}\times \mathcal{A}\times \mathcal{C}\rightarrow \mathcal{Y}$ satisfies community fairness.

The condition for community fairness in Definition \ref{def:CF} is equivalent to:

\begin{eqnarray*}
\forall k\in \mathcal{C}, {\rm Pr}(\widetilde{Y}_{\widehat{Y}, \mathbf{z}}\neq Y|C=k)=\frac{1}{K} \sum_{c=1}^{K}  {\rm Pr}(\widetilde{Y}_{\widehat{Y}, \mathbf{z}} \neq Y|C=c)
\end{eqnarray*}

The error rate of community $k$: ${\rm Pr}(\widetilde{Y}_{\widehat{Y}, \mathbf{z}} \neq Y |C=k)$ can be extended as:
\begin{equation}\label{eq:error_rate}
    \begin{split}
    {\rm Pr} (\widetilde{Y}_{\widehat{Y}, \mathbf{z}} \neq Y |C=k)
            &=\sum_{a=0}^{1}({\rm Pr}_D(\widetilde{Y}_{\widehat{Y}, \mathbf{z}}=0,Y=1,A=a| C=k)+{\rm Pr}_D(\widetilde{Y}_{\widehat{Y}, \mathbf{z}}=1,Y=0,A=a| C=k))\\
            &= \sum_{a=0}^{1} [{\rm Pr}_D(\widehat{Y}=0, Y=1, A=a |C=k)\cdot {\rm Pr}_D(\widetilde{Y}_{\widehat{Y}, \mathbf{z}}=\widehat{Y}|\widehat{Y}=0,A=a, C=k)\\
            &+{\rm Pr}_D(\widehat{Y}=1, Y=1, A=a| C=k)\cdot {\rm Pr}_D(\widetilde{Y}_{\widehat{Y}, \mathbf{z}} \neq \widehat{Y}|\widehat{Y}=1, A=a, C=k)\\
            &+{\rm Pr}_D(\widehat{Y}=1, Y=0, A=a| C=k)\cdot {\rm Pr}_D(\widetilde{Y}_{\widehat{Y}, \mathbf{z}}=\widehat{Y}|\widehat{Y}=1, A=a, C=k)\\
            &+{\rm Pr}_D(\widehat{Y}=0, Y=0, A=a| C=k)\cdot {\rm Pr}_D(\widetilde{Y}_{\widehat{Y}, \mathbf{z}} \neq \widehat{Y}|\widehat{Y}=0, A=a, C=k)]\\
            &=\sum_{a=0}^{1}({\rm FN}^{ak}\cdot z_0 ^{ak}+{\rm TP}^{ak} \cdot (1-z_1 ^{ak})+ {\rm FP}^{ak}\cdot z_1 ^{ak} + {\rm TN}^{ak} \cdot (1-z_0 ^{ak}))/ {\rm Pr} (C=k)\\
            &= \sum_{a=0}^{1}(({\rm FN}^{ak}-{\rm TN}^{ak})\cdot z_0 ^{ak}+ ({\rm FP}^{ak}-{\rm TP }^{ak}) \cdot z_1^{ak}+({\rm TP}^{ak}+ {\rm TN}^{ak}))\cdot \frac{1}{p_k}\\
            &=  \mathbf{n}_k^T \cdot \mathbf{z}_k+ b_k
    \end{split}
    \end{equation}
The last $n$ linear equations of $\mathbf{A}\mathbf{z}=\mathbf{b}$ in (\ref{eq: lp}) are:

for $k=1,2,3,\cdots K$:
\begin{equation}\label{eq:n_lq}
\begin{split}
    0&=-\frac{K-1}{K}\mathbf{n}_k^T \mathbf{z}_k+\frac{1}{K}\sum_{(c\in \mathcal{C},c\neq k)}   \mathbf{n}_c^T\mathbf{z}_c- \frac{1}{K}\sum_{c=1}^{K}(b_k-b_c)\\
    &=-(\mathbf{n}_k^T\mathbf{z}_k+b_k)+\frac{1}{K}\sum_{c \in \mathcal{C}}(\mathbf{n}_c^T\mathbf{z}_c^T+b_c)\\
    &= -{\rm Pr}(\widetilde{Y}_{\widehat{Y}, \mathbf{z}}\neq Y|C=k)+\frac{1}{K}\sum_{c=0}^{K} {\rm Pr} (\widetilde{Y}_{\widehat{Y}, \mathbf{z}} \neq Y| C=c)
\end{split}   
\end{equation}
The last equation is from (\ref{eq:error_rate}).

We can see from (\ref{eq:n_lq}) the last $K$ linear equations of $\mathbf{A}\mathbf{z}=\mathbf{b}$ , the outcome predictor satisfies community fairness.

From the proceeding, if $\mathbf{z}\in \mathbbm{R}^{4K}$ is the solution of the linear program (\ref{eq: lp}), the outcome predictor that satisfies (\ref{eq: variable}) is a fair outcome predictor. The outcome predictor $\widetilde{Y}_{\widehat{Y}, \mathbf{z}}$ that takes values of (\ref{fair_outcome}) has:
\begin{equation*}
    \begin{split}
        {\rm Pr}_D (\widetilde{Y}_{\widehat{Y}, \mathbf{z}}=\widehat{Y}|\widehat{Y}=1, A=a, C=c) &= z_1^{ac}\\
        {\rm Pr}_D (\widetilde{Y}_{\widehat{Y}, \mathbf{z}}=\widehat{Y}|\widehat{Y}=0, A=a, C=c) &= z_0^{ac}
    \end{split}
\end{equation*}

which satisfies (\ref{fair_outcome}).

Thus, the outcome predictor $\widetilde{Y}_{\widehat{Y}, \mathbf{z}}$ is a fair outcome predictor w.r.t. both equal opportunity and community fairness.
 $\diamondsuit$

 \subsection{Proof of Theorem \ref{theo: existence}} \label{App:proof_theo_3.4}
We show that the linear program (\ref{eq: standard_lp}) always has solutions. Before presenting the proof, we first present Farkas' lemma \cite{GoldmanTucker+1957+19+40}:

 Let $\mathbf{A} \in \mathbb{R}^{m \times n}$ and $\mathbf{b} \in \mathbb{R}^m$. Then exactly one of the following two assertions is true:
 
\begin{enumerate}
    \item There exists a $\mathbf{z} \in \mathbb{R}^n$ such that $\mathbf{Az} =\mathbf{ b}$ and $\mathbf{z} \geq 0$.
    \item There exists a $\mathbf{y} \in \mathbb{R}^m$ such that $\mathbf{A}^T \mathbf{y} \geq 0$ and $\mathbf{b}^T \mathbf{y} < 0$.
\end{enumerate}

Faraks' lemma states that either the system $\mathbf{Az}=\mathbf{b}$ has a non-negative solution or the system $\mathbf{A}^T\mathbf{y}\geq 0$ has a solution with $\mathbf{b}^T\mathbf{y}<0$ but not both. Thus, we can show the linear program (\ref{eq: standard_lp}) always exist a solution by showing that the set $\{\mathbf{y}|\mathbf{y}\in \mathbb{R}^{5K+1}, \mathbf{\Bar{A} }^T \mathbf{y}\geq 0, \mathbf{\Bar{b}}^T \mathbf{y}<0\}$ is always empty.

Let: $\mathbf{y}^T= \left[
\begin{array}{cc}
    \mathbf{y}_1^T & \mathbf{y}_2^T 
\end{array}\right]$, with $\mathbf{y}_1\in \mathbb{R}^{K+1}$, $\mathbf{y}_2 \in \mathbb{R}^{4K}$, then, the condition $\mathbf{\Bar{A}}^T\mathbf{y}\geq 0$ is:
\begin{equation}\label{eq:Ay}
    \begin{split}
    \mathbf{\Bar{A}}^T \mathbf{y}=&\left[\begin{array}{cc}
        \mathbf{A}^T &\mathbf{I}  \\
         \mathbf{0}& \mathbf{I} 
    \end{array}
    \right] \cdot \left[\begin{array}{cc}
         \mathbf{y_1} \\
        \mathbf{y_2}
    \end{array}
    \right]\\
    &=\left[\begin{array}{cc}
       \mathbf{A}^T\mathbf{y_1}+\mathbf{I}\mathbf{y_2} \\
        \mathbf{y_2} 
    \end{array}
    \right] \geq 0\\
    \rightarrow \quad & \mathbf{1}_{4K}^T \cdot (\mathbf{A}^T\mathbf{y_1}+\mathbf{y_2})\geq 0\\
    &\mathbf{1}_{4K}^T \cdot \mathbf{y_2}\geq 0
    \end{split}
\end{equation}

The condition $\mathbf{\Bar{b}}^T\mathbf{y}< 0$ is:
\begin{equation}\label{eq:by}
    \begin{split} 
    \mathbf{\Bar{b}}^T \mathbf{y}&= \left[
    \begin{array}{cc}
         \mathbf{b}^T&\mathbf{1}_{4K}^T
    \end{array}\right] \cdot \left[\begin{array}{cc}
        \mathbf{y_1}  \\
         \mathbf{y_2} 
    \end{array}\right]\\
    &= \mathbf{b}^T\mathbf{y_1}+\mathbf{1}_{4K}^T \mathbf{y_2}\\
    &= \frac{1}{2}\mathbf{A}^T \mathbf{y_1}+\mathbf{1}_{4K}^T \mathbf{y_2}<0
    \end{split}
\end{equation}
The last equation above is from the fact: $\frac{1}{p_c}({\rm TN}^{0c}+{\rm TP}^{0c}+{\rm TN}^{1c}+{\rm TP}^{1c})= 1-\frac{1}{p_c}({\rm FN}^{0c}+{\rm FP}^{0c}+{\rm FN}^{1c}+{\rm FP}^{1c})$.

When $\mathbf{1}_{4K}^T\mathbf{y_2}\geq 0 $, which is the second condition in (\ref{eq:Ay}), the first condition in (\ref{eq:Ay}): $\mathbf{1}_{4K}^T \cdot (\mathbf{A}^T\mathbf{y_1})\geq -\mathbf{1}_{4K}^T\mathbf{y_2}$ is always conflict with  the condition (\ref{eq:by}): $\mathbf{1}_{4K}^T \cdot (\mathbf{A}^T\mathbf{y_1})< -2\cdot \mathbf{1}_{4K}^T\mathbf{y_2}$, as $-2\cdot \mathbf{1}_{4K}^T\mathbf{y_2}\leq -\mathbf{1}_{4K}^T\mathbf{y_2} $.

Thus, the set: $\{\mathbf{y}|\mathbf{y}\in \mathbb{R}^{5K+1}, \mathbf{\Bar{A} }^T \mathbf{y}\geq 0, \mathbf{\Bar{b}}^T \mathbf{y}<0\}$ is always empty. This indicates the system $\mathbf{\Bar{A} \Bar{z}}=\mathbf{\Bar{b}}$ always has non-negative solutions. The variables in the linear program represent probabilities that are bounded in $[0, 1]$. Therefore,  the objective function of  (\ref{eq: standard_lp}) is bounded. The linear program always has solutions.

 $\diamondsuit$

 \subsection{Proof of Theorem \ref{theo:fair_outcome_predictor}}\label{App: proof_theo_3.5}
 \vspace{0.1in}
{\bf Proof:} The ${\rm Pr}_D (\widetilde{Y}\neq Y)$ in first condition can be extended as:
\begin{equation}\label{eq:error_rate_extension}
    \begin{split}
    {\rm Pr}_D (\widetilde{Y} \neq Y)
            &=\sum_{c=1}^{K}\sum_{a=0}^{1}({\rm Pr}_D(\widetilde{Y}=0,Y=1,A=a, C=k)+{\rm Pr}_D(\widetilde{Y}=1,Y=0,A=a, C=k))\\
            &= \sum_{c=1}^{K}\sum_{a=0}^{1} [{\rm Pr}_D(\widehat{Y}=0, Y=1, A=a, C=k)\cdot {\rm Pr}_D(\widetilde{Y}=\widehat{Y}|\widehat{Y}=0,A=a, C=k)\\
            &+{\rm Pr}_D(\widehat{Y}=1, Y=1, A=a, C=k)\cdot {\rm Pr}_D(\widetilde{Y} \neq \widehat{Y}|\widehat{Y}=1, A=a, C=k)\\
            &+{\rm Pr}_D(\widehat{Y}=1, Y=0, A=a, C=k)\cdot {\rm Pr}_D(\widetilde{Y}=\widehat{Y}|\widehat{Y}=1, A=a, C=k)\\
            &+{\rm Pr}_D(\widehat{Y}=0, Y=0, A=a, C=k)\cdot {\rm Pr}_D(\widetilde{Y} \neq \widehat{Y}|\widehat{Y}=0, A=a, C=k)]\\
            &=\sum_{c=1}^{K}\sum_{a=0}^{1}({\rm FN}^{ak}\cdot z_0 ^{ak}+{\rm TP}^{ak} \cdot (1-z_1 ^{ak})+ {\rm FP}^{ak}\cdot z_1 ^{ak} + {\rm TN}^{ak} \cdot (1-z_0 ^{ak}))\\
            &= \mathbf{c}^T \mathbf{z}+\sum_{c=0}^{K}b_c p_c\\
            &= \mathbf{\Bar{c}}^T \mathbf{\Bar{z}}+\sum_{c=0}^{K}b_c p_c
    \end{split}
    \end{equation}
    As we show in proposition (\ref{prop:lp}), the constraints of equal opportunity and community fairness in the linear program (\ref{eq: lp}) (or a standard form (\ref{eq: standard_lp})) are:
    \begin{equation*}
        \begin{split}
            &\mathbf{\Bar{A}} \mathbf{\Bar{z}}=\mathbf{\Bar{b}}\\
            \Rightarrow \quad & 
            \mathbf{\Bar{A}}^T\mathbf{\Bar{A}}\mathbf{\Bar{z}}=\mathbf{\Bar{A}}^T\mathbf{\Bar{b}}
        \end{split}
    \end{equation*}
For matrix $\mathbf{\Bar{A}}^T\mathbf{\Bar{A}}$, we have:
\begin{equation}
    \begin{split}
        \lambda_{\rm min}(\mathbf{\Bar{A}}^T\mathbf{\Bar{A}})\mathbf{\Bar{z}}^T\mathbf{\Bar{z}}\leq \mathbf{\Bar{z}}^T\mathbf{\Bar{A}}^T \mathbf{\Bar{A}}\mathbf{\Bar{z}}
    \end{split}
\end{equation} 
where, $\lambda_{\rm min}(\mathbf{\Bar{A}}^T\mathbf{\Bar{A}})$ is the smallest eigenvalue of $\mathbf{\Bar{A}}^T \mathbf{\Bar{A}}$.

Then, we show the $l_2$ norm of $\mathbf{\Bar{z}}$ is bounded:
\begin{equation}\label{eq:lower_bound}
   \begin{split}
              &\lambda_{\rm min}(\mathbf{\Bar{A}}^T\mathbf{\Bar{A}})\mathbf{\Bar{z}}^T\mathbf{\Bar{z}}\leq \mathbf{\Bar{z}}^T\mathbf{\Bar{A}}^T \mathbf{\Bar{A}}\mathbf{\Bar{z}}\leq \|\mathbf{\Bar{z}}\|_2 \|\mathbf{\Bar{A}}^T \mathbf{\Bar{A}}\mathbf{\Bar{z}}\|_2=\|\mathbf{\Bar{z}}\|_2 \|\mathbf{\Bar{A}}^T \mathbf{\Bar{b}}\|_2\\
              \Rightarrow \quad & \lambda_{\rm min}(\mathbf{\Bar{A}}^T\mathbf{\Bar{A}}) \|\mathbf{\Bar{z}}\|_2\leq \|\mathbf{\Bar{A}}^T\mathbf{\Bar{b}}\|_2\\
              \Rightarrow \quad &\|\mathbf{z}\|_2\leq\frac {\|\mathbf{\Bar{A}}^T\mathbf{\Bar{b}}\|_2}{\underline{\sigma}^2}
   \end{split} 
\end{equation}
where, $\underline{\sigma}$ is the smallest singular value of the matrix $\mathbf{\Bar{A}}$.

Thus, the ${\rm Pr}_D (\widetilde{Y}\neq Y)$ has: 
\begin{equation}\label{eq:bound}
    \begin{split}
        {\rm Pr}_D (\widetilde{Y}\neq Y)&= \mathbf{\Bar{c}}^T \mathbf{\Bar{z}}+\sum_{c=0}^{K}b_c p_c\\
        &\geq -\|\mathbf{\Bar{c}}\|_\infty \|\mathbf{\Bar{z}}\|_\infty+\sum_{c=0}^{K}b_cp_c \quad \textit{(all elements in c are negative)} \\
        &\geq -\|\mathbf{\Bar{c}}\|_\infty \|\mathbf{\Bar{z}}\|_2+\sum_{c=0}^{K}b_c p_c\quad (\|\mathbf{\Bar{z}}\|_2 \geq \|\mathbf{\Bar{z}}\|_\infty) \\
        &\geq -\|\mathbf{\Bar{c}}\|_\infty \frac {\|\mathbf{\Bar{A}}^T\mathbf{\Bar{b}}\|_2}{\underline{\sigma}^2}+\sum_{c=0}^{K}b_c p_c\quad
        \textit{(the upper bound (\ref{eq:lower_bound}}))\\
    \end{split}
\end{equation}
with $b_c p_c = ({\rm TN}^{0c}+{\rm TP}^{0c}+{\rm TN} ^{1c}+{\rm TP}^{1c})$.

When $\Delta < -\|\mathbf{\Bar{c}}\|_\infty \frac {\|\mathbf{\Bar{A}}^T\mathbf{\Bar{b}}\|_2}{\underline{\sigma}^2}+\sum_{c=0}^{K}b_c p_c$, the inequality: ${\rm Pr}_D (\widetilde{Y}\neq Y)\geq -\|\mathbf{\Bar{c}}\|_\infty \frac {\|\mathbf{\Bar{A}}^T\mathbf{\Bar{b}}\|_2}{\underline{\sigma}^2}+\sum_{c=0}^{K}b_c p_c$ is conflict with the first condition in (\ref{eq: epsilon_fair_predictor}): ${\rm Pr}_D (\widetilde{Y}\neq Y)\leq \Delta$.

Thus, if $\Delta < -\|\mathbf{\Bar{c}}\|_\infty \frac {\|\mathbf{\Bar{A}}^T\mathbf{\Bar{b}}\|_2}{\underline{\sigma}^2}+\sum_{c=0}^{K}b_c p_c $, the fairness condition and $\epsilon$-accurate condition are incompatible with each other.
$\diamondsuit$
\subsection{Proof of Theorem \ref{Theorem: accuracy_loss}} \label{App: Proof_of_theo_3.6}
The error rate of the fair outcome predictor is demonstrated in (\ref{eq:error_rate_extension}), which is: ${\rm Pr}_D (\widetilde{Y}\neq Y)= \mathbf{c}^T \mathbf{z}+\sum_{c=0}^{K}b_c\cdot p_c$.

The predictor $\widetilde{Y}$ is optimal when $\mathbf{z}=\mathbf{1}_{4k}$, the error rate of the optimal predictor is: ${\rm Pr}_D (\widehat{Y}\neq Y)= \mathbf{c}^T \mathbf{1}_{4K}+\sum_{c=0}^{K}b_c\cdot p_c$.

Thus, the minimum error we need to compromise for enforcing group fairness and community fairness is:

${\rm Pr}_D (\widetilde{Y}\neq Y)-{\rm Pr}_D (\widehat{Y}\neq Y)=\mathbf{c}^T \mathbf{z}-\mathbf{c}^T \mathbf{1}_{4K}$.
 $\diamondsuit$
 
\section{Additional Experimental Details and Results}




\subsection{Models and Hyperparameters}
\label{App: exp_details}
We provide the models and hyperparameters of post-FFL for each dataset.
We run the experiments on  on our local Linux server with a 16-Core 4.00 GHz AMD RYZEN Threadripper Pro 5955WX Processor. We implement all code in TensorFlow \cite{tensorflow2015-whitepaper}, simulating a federated network with one server and several local communities.

For Adult, in each local community, we randomly divide the data into three subsets: 60\% for the training set, 20\% for the validation set, and 20\% for the test set. We first implement the \textit{FedAvg} algorithm. For each communication round in \textit{FedAvg}, the number of participating communities is set to \(N=2\). We set the number of local update epochs to \(E=1\) with a batch size of \(B=512\). The local models are logistic regression classifiers with two layers, containing 64 and 32 nodes, respectively. We use \textit{Relu} as the activation functions for each hidden layers. These models are trained using the Adam optimizer with a learning rate of \( \eta=0.001 \). We select the number of rounds that minimizes the  validation loss, and then report the evaluation metrics on the test dataset. We construct a linear program using the training data. Finally, we apply post-processing to the test dataset based on the solution from the linear program and report its evaluation metrics.

For the Diabetes dataset, we similarly split the local dataset into 60\% for training, 20\% for validation, and 20\% for testing. The number of participating communities for \textit{FedAvg} is set to \(N=7\). We maintain the number of local update epochs at \(E=1\) with a batch size of \(B=256\). We follow the same model structure, optimization algorithm, and evaluation process as with the UCI Adult dataset and report its evaluation metrics.

\begin{figure*}[ht]
\begin{center}
\centerline{\includegraphics[width=2 in]{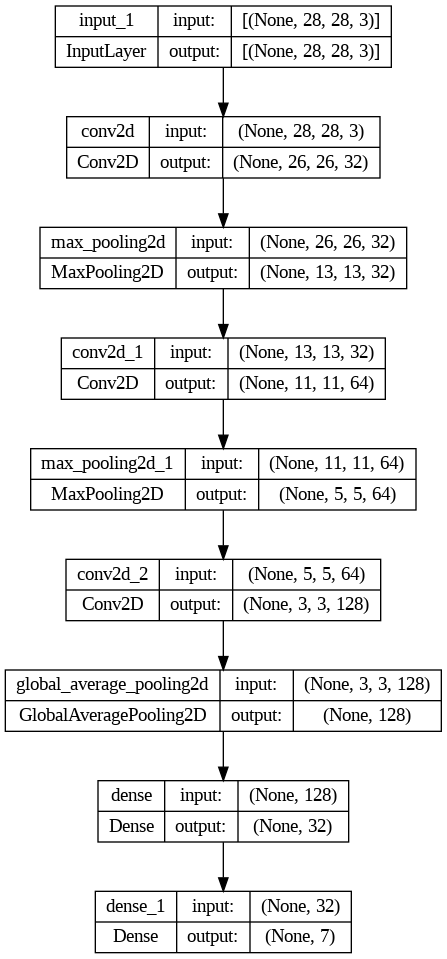}}
\caption{The CNN structure used for HM10000 Dataset}
\label{fig:CNN_model}
\end{center}
\vskip -0.2in
\end{figure*}

For HM 10000, we similarly split the local dataset into 60\% for training, 20\% for validation, and 20\% for testing. We resize the the images to $28\times28 \times 3$. We set the number of local update epochs to \(E=5\) with a batch size of \(B=32\). The local models' structure are as shown in Fig. \ref{fig:CNN_model} The loss function we use for multi-class classification task is sparse categorical cross entropy. The CNN models are trained using the Adam optimizer with a learning rate \(\eta= 0.001\). We select the number of rounds that minimizes the  validation loss, and then report the evaluation metrics on the test dataset. We construct a linear program using the training data. Finally, we apply post-processing to the test dataset based on the solution from the linear program and report its evaluation metrics.

\subsection{Baselines}\label{app: baselines}
\begin{figure*}[h]
      \caption{Adult dataset: Training curves of \textit{Fedavg} (left), \textit{FairFed} (middle), and \textit{q-FedAvg} (right)}
  \centering
  \includegraphics[width=1.1\textwidth]{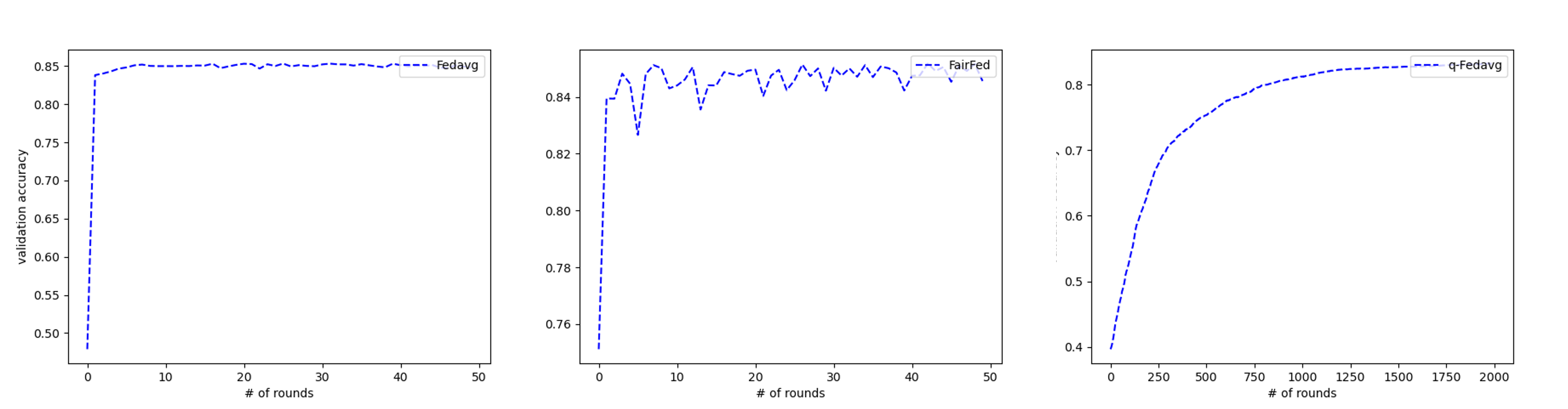}
  \label{fig:convergence}
\end{figure*}

In Section \ref{sec:compare with other algorithms}, we use \textit{q-FedAvg} \cite{li2019fair} and AFL \cite{mohri2019agnostic} as baselines for community fairness, and \textit{FairFed} \cite{ezzeldin2023fairfed} as the baseline for group fairness. Detailed descriptions and convergence curves of all baselines are provided below:
\begin{itemize}
    \item \textit{q-FedAvg} \cite{li2019fair} improves community fairness in Federated Learning (FL) by minimizing an aggregated reweighed loss, parameterized by $q$. The algorithm assigns greater weight to devices with higher loss. The parameter $q$ controls the trade-off between community fairness and model utility. In our experiments, we set $q=4$ for Adult, following the recommendation in the original implementation of the \textit{q-FedAvg} paper and $q=1$ for HM1000.
    
    \item \textit{FairFed}, \cite {ezzeldin2023fairfed} improves equal opportunity in FL, which also minimizes an aggregated reweighed loss, parameterized by $\beta$. The weights are a function of the mismatch between the global EOD (on the full dataset) and the local EOD at each community, favoring communities whose local measurement match the global measurements. $\beta$ is the parameter that control the tradeoff between the group fairness and model utility. We follows the initial paper's setting: $\beta=1$ for Adult and $\beta=2$ for HM1000.
\end{itemize}
The convergence curves of \textit{FedAvg}, \textit{q-FedAvg}, and \textit{FairFed} are shown in Fig. \ref{fig:convergence}.

\subsection{Additional Results of Different Data Heterogeneity Level }
\label{app:additional results on HM1000}
We provide results of \textit{HM10000} with various degrees of data heterogeneity in Table \ref{Table: data hetergeneity 2}. Similar as \textit{Adult}, we introduce community heterogeneity by adjusting the mixture of individuals with specific target and sensitive attributes.

We define class 1 as unprotected Melanocytic Nevi samples $(y = 4, s = 1)$, class 2 as protected Melanocytic Nevi  $(y = 4, s = 0)$, class 3 as unprotected non-Melanocytic Nevi  $(y\neq 0, s =
1)$, and class 4 as protected non-Melanocytic Nevi  $(y\neq4, s = 0)$. We sample the local communities’ datasets so that the class proportion of
each community is as shown in Fig. 5.3. For scenario 1, Data is i.i.d. distributed across all communities. The sensitive
attribute is totally skewed across two communities in scenario 2. Both the target and sensitive attributes are imbalanced
across communities in scenario 3
\begin{figure*}[h]
\subfigure[Scenario 1 (s1)]{
\begin{minipage}[t]{0.35\linewidth}
\centering
\includegraphics[width=2.6in]{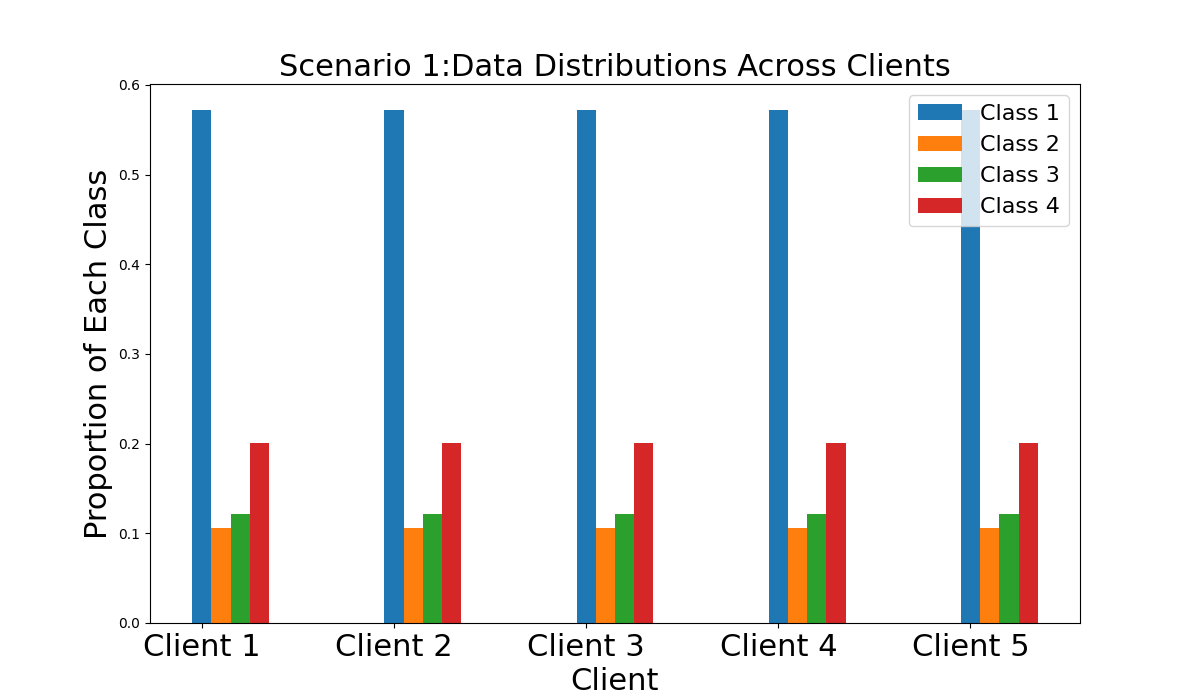}
\end{minipage}%
}%
\subfigure[Scenario 2 (s2)]{
\begin{minipage}[t]{0.35\linewidth}
\centering
\includegraphics[width=2.6in]{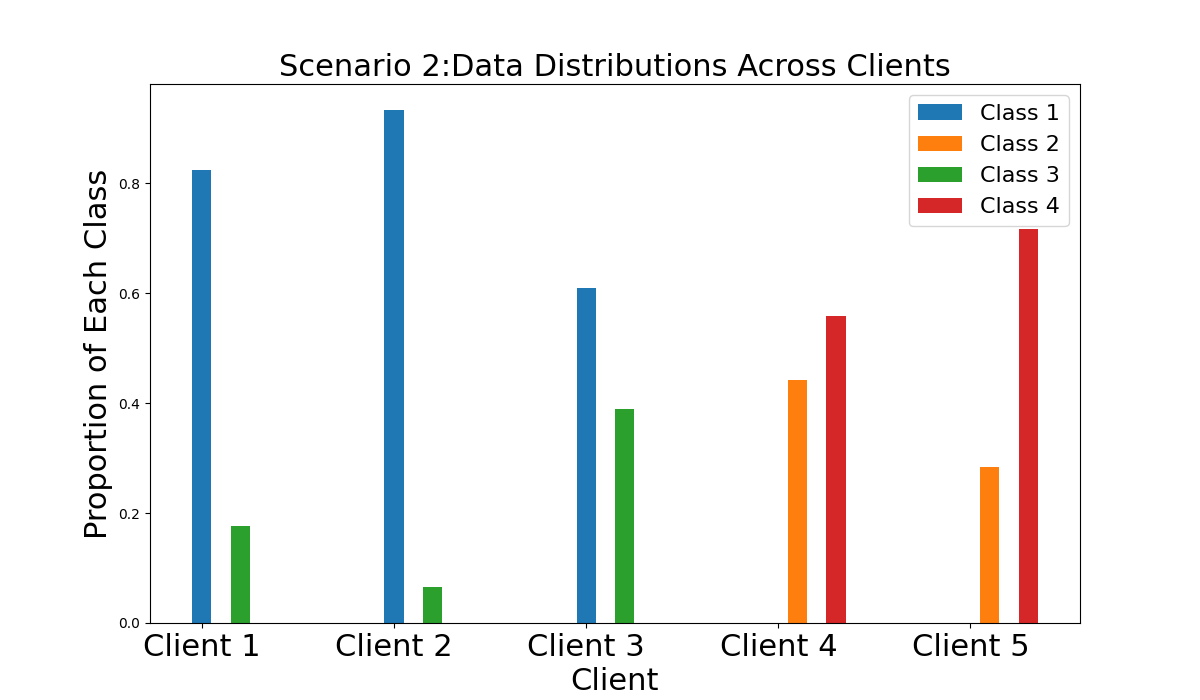}
\end{minipage}%
}%
\subfigure[Scenario 3 (s3)]{
\begin{minipage}[t]{0.35\linewidth}
\centering
\includegraphics[width=2.6in]{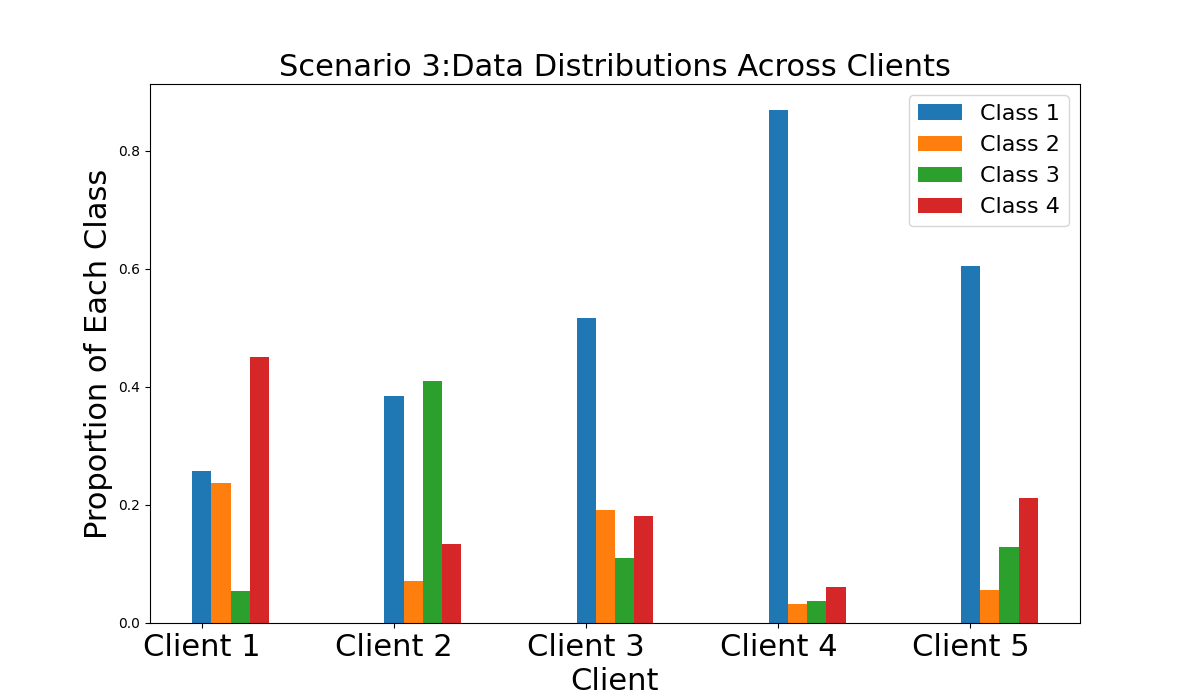}
\end{minipage}%
}%
\caption{HM1000: The class proportion of the three scenarios}
\end{figure*}

\begin{table}[ht]
    \centering
    \small
    \caption{Experimental results with different degrees of heterogeneity(HM1000)}
    \label{Table: data_hetergeneity_2}
    \begin{tabular}{c|c|cccccccc|c}
        \hline
        Scenario&Algorithms&Avg-Acc&EOD&Acc@1&Acc@2&Acc@3&Acc@4&Acc@5&AD&Accuracy trade for fairness\\ \hline
        s1& FedAvg&0.695&0.002&0.694&0.693&0.714&0.693&0.680&0.034&-\\ \hline s1&PostFFL&0.688&0.007&0.699&0.682&0.719&0.680&0.664&0.056&0.007\\ \hline s2&FedAvg&0.675&0.076&0.865&0.932&0.637&0.447&0.270&0.662&-\\ \hline
        s2&Post-FFL&0.330&0.00&0.344&0.317&0.318&0.346&0.334&0.029&0.345 \\ \hline s3&FedAvg&0.681&0.060&0.522&0.500&0.697&0.900&0.658&0.451&-\\ \hline
        s3&Post-FFL&0.509&0.016&0.533&0.498&0.516&0.513&0.476&0.058&0.172\\ \hline       
    \end{tabular}
\end{table}


\end{document}